\documentclass[10pt,twocolumn,letterpaper]{article}

\usepackage[pagenumbers]{iccv} 

\definecolor{iccvblue}{rgb}{0.21,0.49,0.74}
\usepackage[pagebackref,breaklinks,colorlinks,allcolors=iccvblue]{hyperref}
\usepackage[ruled, lined, linesnumbered, commentsnumbered, longend]{algorithm2e}
\usepackage{algpseudocode}
\usepackage{adjustbox}
\usepackage{multirow}
\usepackage{colortbl, xcolor}
\usepackage{makecell}

\def\paperID{1649} 
\def\confName{ICCV}
\def\confYear{2025}

\title{AccVideo: Accelerating Video Diffusion Model with Synthetic Dataset}

\author{%
  Haiyu Zhang\textsuperscript{1,2}\footnotemark[1],\ \ 
  Xinyuan Chen\textsuperscript{2},\ \ 
  Yaohui Wang\textsuperscript{2},\ \ 
  Xihui Liu\textsuperscript{3},\ \ 
  {Yunhong Wang\textsuperscript{1}},\ \ 
  {Yu Qiao\textsuperscript{2}\footnotemark[2]}\\
    {\textsuperscript{1}Beihang University \ \ 
    \textsuperscript{2}Shanghai AI Laboratory \ \ 
    \textsuperscript{3}The University of Hong Kong
    }\\
    \small{\textsuperscript{1}\texttt{\{zhyzhy,yhwang\}@buaa.edu.cn}} \ \ 
 \small{\textsuperscript{2}\texttt{\{chenxinyuan,wangyaohui,qiaoyu\}@pjlab.org.cn}}\\
  \small{\textsuperscript{3}\texttt{xihuiliu@eee.hku.hk}}\\
    \small\url{https://aejion.github.io/accvideo} \\
}

\begin{document}

\twocolumn[{%
 \renewcommand\twocolumn[1][]{#1}%
 \maketitle
 \centering
 \vspace{-6mm}
 \includegraphics[width=\textwidth, trim={0 0.5cm 0 0.5cm}]{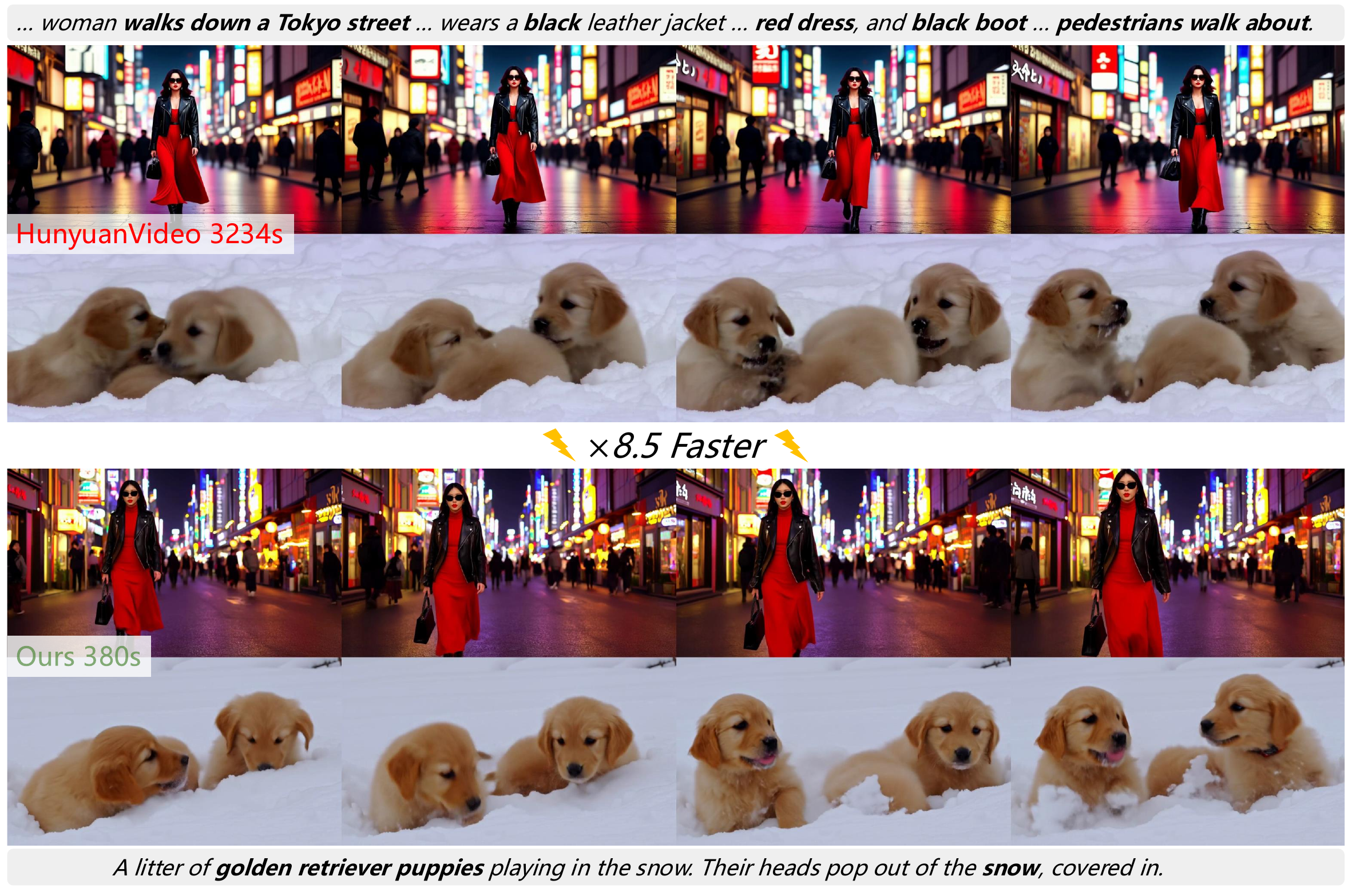}
 \captionof{figure}{ 
    Video diffusion models can generate high-quality videos, but they require dozens of inference steps, resulting in slow generation process. For instance, HunyuanVideo~\cite{kong2024hunyuanvideo} takes 3234 seconds to generate a 5-seconds, 720$\times$1280, 24fps video on a single A100 GPU. In contrast, our method accelerates video diffusion models through distillation, achieving 8.5$\times$ improvements in generation speed while maintaining comparable quality.
    \label{fig:teaser}
   }
 \vspace{1.8em}
}]

\renewcommand{\thefootnote}{\fnsymbol{footnote}}

\footnotetext[1]{Work done when Haiyu Zhang interned at Shanghai AI Laboratory.}
\footnotetext[2]{Corresponding author}

\maketitle

\begin{abstract}
Diffusion models have achieved remarkable progress in the field of video generation. However, their iterative denoising nature requires a large number of inference steps to generate a video, which is slow and computationally expensive. In this paper, we begin with a detailed analysis of the challenges present in existing diffusion distillation methods and propose a novel efficient method, namely \textbf{AccVideo}, to reduce the inference steps for accelerating video diffusion models with synthetic dataset. We leverage the pretrained video diffusion model to generate multiple valid denoising trajectories as our synthetic dataset, which eliminates the use of useless data points during distillation. Based on the synthetic dataset, we design a trajectory-based few-step guidance that utilizes key data points from the denoising trajectories to learn the noise-to-video mapping, enabling video generation in fewer steps. Furthermore, since the synthetic dataset captures the data distribution at each diffusion timestep, we introduce an adversarial training strategy to align the output distribution of the student model with that of our synthetic dataset, thereby enhancing the video quality. Extensive experiments demonstrate that our model achieves 8.5$\times$ improvements in generation speed compared to the teacher model while maintaining comparable performance. Compared to previous accelerating methods, our approach is capable of generating videos with higher quality and resolution, \ie, 5-seconds, 720$\times$1280, 24fps.

\end{abstract}    

\section{Introduction}
\label{sec:intro}

Video generation has garnered significant attention due to its ability to simulate the real physical world \cite{agarwal2025cosmos, sora}, as well as its promising applications in entertainment, such as content creation \cite{ma2024latte,wang2024lavie,kong2024hunyuanvideo,yang2024cogvideox,chen2024videocrafter2,zheng2024open,guo2023animatediff,jin2024pyramidal,blattmann2023align}, filmmaking \cite{kling, sora}, video games \cite{yang2024playable,valevski2024diffusion}, and customized media generation \cite{ma2024follow, hu2024animate,lin2025omnihuman,kling}. 

With advancements in data curation pipeline \cite{kong2024hunyuanvideo, agarwal2025cosmos} and scalable model architectures \cite{peebles2023scalable}, diffusion models \cite{karras2022elucidating,ho2020denoising} and flow matching~\cite{liu2022flow,lipman2022flow} have emerged as widely used frameworks in video generation, owing to their impressive generative capabilities. However, video diffusion models require iterative denoising of Gaussian noise to generate the final videos, making them typically demand dozens of inference steps. This process is both time-consuming and computationally intensive. For instance, HunyuanVideo~\cite{kong2024hunyuanvideo} requires 3234s to generate a 5s, 720$\times$1280, 24fps video on a single NVIDIA A100 GPU, as shown in Fig.~\ref{fig:teaser}.

Recently, significant progress has been made in the field of accelerating image diffusion models by distillation~\cite{salimans2022progressive, frans2024one, berthelot2023tract, yan2024perflow, yin2024one, yin2024improved, luo2024one, sauer2024adversarial, wang2022diffusion, xu2024ufogen, lin2024sdxl, sauer2024fast}. However, distilling video diffusion models remains a challenge that requires further exploration. Although distillation methods for image diffusion models can be extended to video diffusion models, they require a large amount of data and computational resources~\cite{lin2025diffusion, yin2024slow} due to the useless data points, which do not lie on the denoising trajectories of the teacher model. Moreover, the useless data points cause the teacher model to offer unreliable guidance for the student model, adversely affecting the video quality. The spatial-temporal complexity of videos exacerbates these issues.

To be concrete, the useless data points during distillation are caused by dataset mismatching or Gaussian noise mismatching. Dataset mismatching refers to the inconsistency between the dataset used for training the teacher model and the dataset utilized during the distillation process, which arises due to the difficulty in accessing the training dataset of the teacher model. Gaussian noise mismatching refers to the misalignment between Gaussian noise and the data, a phenomenon that occurs during the forward diffusion operation, also called the flow operation.

In this paper, we begin with a detailed analysis of the challenges present in existing diffusion distillation models. Based on the analysis, we try to avoid the use of useless data points during the distillation process and propose a novel efficient distillation method, namely \textbf{AccVideo}, which aims to accelerate video diffusion models with synthetic dataset. Specifically, we first present a synthetic dataset, SynVid, comprising 110K denoising trajectories and high-quality videos generated by HunyuanVideo~\cite{kong2024hunyuanvideo} with fine-grained text prompts. The data points on the denoising trajectories are intermediate results leading to the correct output, making them all valid and meaningful. Then, we design a trajectory-based few-step guidance that selects a few of data points from each denoising trajectory to construct a shorter noise-to-video mapping path, enabling the student model to generate videos in fewer steps. To further exploit the data distribution captured by our synthetic dataset, we propose an adversarial training strategy to align the output distribution of the student model with that of our synthetic dataset at each diffusion timestep, thereby enhancing the quality of generated videos. It is noteworthy that the data points used in our method deliver precise guidance for the student model, markedly enhancing training efficiency and reducing the number of data. Our model is trained using only 8 A100 GPUs with 38.4K synthetic data for 12 days, yet it is capable of generating high-quality 5-seconds, 720$\times$1280, 24fps videos. Moreover, we obviate the complex regularization designs for the adversarial training~\cite{lin2025diffusion} by combining the trajectory-based few-step guidance.

To summarize, our contributions are as follows: 
\begin{itemize}
    \item We provide a detailed analysis of existing diffusion distillation models and propose a novel efficient method to accelerate video diffusion models, eliminating the presence of useless data points and enabling efficient distillation.
    \item We present SynVid, a synthetic video dataset containing 110K high-quality synthetic videos, denoising trajectories, and corresponding fine-grained text prompts.
    \item We design a trajectory-based few-step guidance that leverages key data points from the synthetic dataset to learn the noise-to-video mapping with fewer steps and an adversarial training strategy, which effectively utilizes the synthetic dataset, enhancing the performance.
    \item Extensive experiments demonstrate that we achieve 8.5$\times$ improvements in generation speed compared to the teacher model while maintaining comparable performance. Moreover, we produce videos with higher quality and resolution, \ie, 5-seconds, 720$\times$1280, 24fps, compared to previous accelerating methods.
\end{itemize}

\section{Related Work}
\label{sec:related_work}

\textbf{Video Diffusion Models.} With the advancement of large-scale video data and large-scale models, video diffusion models have achieved remarkable success. The pioneering work VDM~\cite{ho2022video} extends the 2D U-Net~\cite{ronneberger2015u} used in image generation to 3D UNet and proposes joint training with both images and videos.~\cite{he2022latent, zhou2022magicvideo, wang2023modelscope, guo2023animatediff, chen2024videocrafter2} adopt the latent diffusion model (LDM)~\cite{rombach2022high} to learn video data distribution in the latent space, significantly reducing computational complexity. To further increase the resolution of generated videos,~\cite{blattmann2023align, ho2022imagen, wang2024lavie} employ cascaded video diffusion models, which decomposes the video generation process into subtasks, such as key frame generation, frame interpolation, and super-resolution. With the impressive video generation capabilities shown by Sora~\cite{sora}, the diffusion transformer architecture (DiT)~\cite{peebles2023scalable} has gradually become the mainstream backbone for video diffusion models. Latte~\cite{ma2024latte} and GenTron~\cite{chen2023gentron} explore different variants of DiT for video generation. Snap Video ~\cite{menapace2024snap}, HunyuanVideo~\cite{kong2024hunyuanvideo}, and Cosmos~\cite{agarwal2025cosmos} replace 1D+2D self-attention with 3D self-attention and leverage more careful data curation pipelines, larger-scale models, and more computational resources, achieving state-of-the-art video generation performance. For more details on video diffusion models, please refer to~\cite{xing2024survey}.

\noindent\textbf{Accelerating Image Diffusion Models.} Here, we introduce distillation techniques to accelerate image diffusion models. \cite{salimans2022progressive, frans2024one, berthelot2023tract} progressively distill the teacher model, enabling the student model to learn the noise-to-image mapping with fewer inference steps. PeRFlow~\cite{yan2024perflow} divides the denoising process into several windows and learns the data mapping within each window. LCM~\cite{luo2023latent} utilize consistency models~\cite{song2023consistency} in the image latent space to accelerate pretrained image diffusion models. Inspired by Variation Score Distillation (VSD)~\cite{wang2024prolificdreamer} and Score Distillation Sampling (SDS)~\cite{poole2022dreamfusion}, ~\cite{yin2024one, yin2024improved, luo2024one, sauer2024adversarial} propose the distribution matching loss, which aligns the real and fake image distributions by utilizing the score function derived from the diffusion model. Additionally, \cite{wang2022diffusion, xu2024ufogen, lin2024sdxl, sauer2024fast} employ adversarial loss to train few-step or one-step image generators, using a diffusion model as the feature extractors. \cite{liu2023instaflow, luhman2021knowledge} share similarities with our approach, they also utilize synthetic data to learn the noise-to-data mapping in fewer steps. However, they overlook the distribution information contained in the denoising trajectory, resulting blurry outputs.

\noindent\textbf{Accelerating Video Diffusion Models.} Recent studies primarily accelerate video diffusion models from three perspectives: efficient model architectures, high-compression-rate Variational Autoencoders (VAEs) \cite{kingma2013auto}, and distillation techniques. LinGen~\cite{wang2024lingen} employs the linear-complexity Mamba2 block~\cite{dao2024transformers}, significantly reducing the training computational costs. LTX-Video~\cite{hacohen2024ltx} utilizes a carefully designed Video-VAE that achieves a high compression ratio, reducing the input dimensions of the diffusion model and enhancing generation speed. AnimateDiff-Lightning~\cite{lin2024animatediff} extends progressive adversarial distillation~\cite{lin2024sdxl} to video diffusion models. T2V-Turbo~\cite{li2024t2v} and T2V-Turbo-V2~\cite{li2024t2v2} leverage refined consistency distillation loss~\cite{luo2023latent} to reduce the number of inference steps and enhance video quality using reward models. Concurrent works, CausVid~\cite{yin2024slow} and APT~\cite{lin2025diffusion}, leverage distribution matching loss and adversarial loss, respectively, to distill video diffusion models. They attempt to address the dataset mismatch by training the teacher model with their own video dataset before distillation. However, it requires substantial computational resources, especially when training models for high-resolution video generation. Moreover, compared to them, our approach is capable of generating higher quality and resolution videos, \ie, 5-seconds, 720$\times$1280, 24fps.
\section{Preliminaries}
\label{sec:Preliminary}
\subsection{Flow Matching}

In this section, we provide a brief overview about Flow Matching~\cite{lipman2022flow}, which is commonly used in generative tasks. Flow Matching transforms a complex data distribution $p_{0}(x)$ into the simple standard normal distribution $p_{1}(x)=\mathcal{N}(0, I)$ through a conditional probability paths $p_t(x \vert x_0)$, where $x_0\sim p_{0}(x)$ and $t\in [0,1]$. In \cite{lipman2022flow}, the conditional probability paths have the form:
\begin{equation}
    p_t(x\vert x_0) = \mathcal{N}(x\, \vert\, \mu_t(x_0), \sigma_t(x_0)^2 I ),
\end{equation}
where $\mu_t(x_0)$ is the time-dependent mean, while $\sigma_t(x_0)$ describes the time-dependent scalar standard deviation (std). According to the Optimal Transport theory, the mean and the std are set to change linearly in time,
\begin{equation}
    \mu_t(x_0) = (1-t)x_0, \sigma_t(x_0) = t.
\end{equation}

Then, the sample $x_t=(1-t)x_0+tx_1\sim p_t(x\vert x_0)$ is obtained through forward diffusion operation. Here, $x_1 \sim p_{1}(x)$. A model $v$ parameterized by $\theta$ is trained to predict the velocity $u_t(x|x_0)=x_1-x_0$, which guides the sample $x_t$ towards the data $x_0$. The training loss has the form:
\begin{equation}
     \mathcal{L}_{\text{FM}}(\theta) 
     =\mathbb{E}_{t, p_0(x), p_1(x)} \Big\|v_{\theta}(x_t,t) - u_t(x|x_0)) \Big\|^2. 
     \label{training}
\end{equation}

After training, a sampled Gaussian noise $x_1\sim p_1(x)$ can be denoised to data $x_0 \sim p_0(x)$ by integrating the predicted velocity $v_{\theta}(x_t,t)$ through the first-order Euler
Ordinary Differential Equation (ODE) solver.
\subsection{Analysis of Previous Distillation Methods}
\label{sec:analysis}

\begin{figure}
  \centering
  \setlength{\abovecaptionskip}{0pt}
  \setlength{\belowcaptionskip}{0pt}
  \includegraphics[width=1.\linewidth]{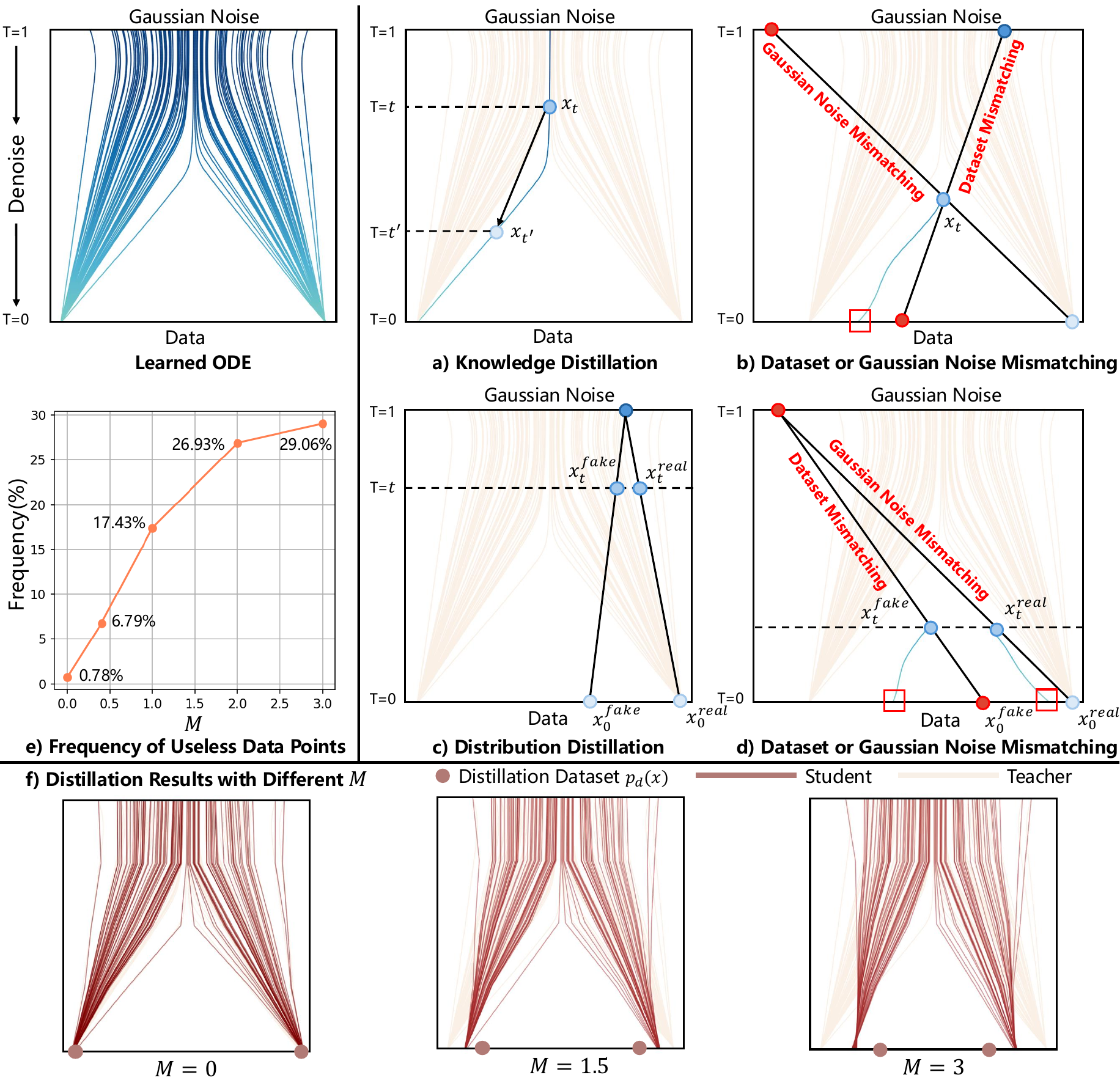}

   \caption{\textbf{1D Toy Experiment.} We employ Flow Matching objective~\cite{lipman2022flow} to train the teacher model, which learns the ODE that maps Gaussian distribution to the data distribution. The data distribution consists of two data points. \textbf{a)} illustrates the knowledge distillation methods, where a student model is trained to mimic the teacher model's denoising process. \textbf{b)} highlights the challenges posed by dataset or Gaussian noise mismatching in knowledge distillation, which can lead to unreliable guidance. \textbf{c)} demonstrates the distribution matching methods, which aims to align the output distribution of the student model with that of the teacher model. \textbf{d)} emphasizes the issue in distribution matching, which can result in inaccurate guidance. \textbf{e)} illustrates the frequency of useless data points in relation to $M$. \textbf{f)} shows the distillation results at various values of $M$.}
   \label{fig:motivation}
   \vspace{-2mm}
\end{figure}

\begin{figure*}
  \centering
  \setlength{\abovecaptionskip}{0pt}
  \setlength{\belowcaptionskip}{0pt}
  \includegraphics[width=1.\linewidth]{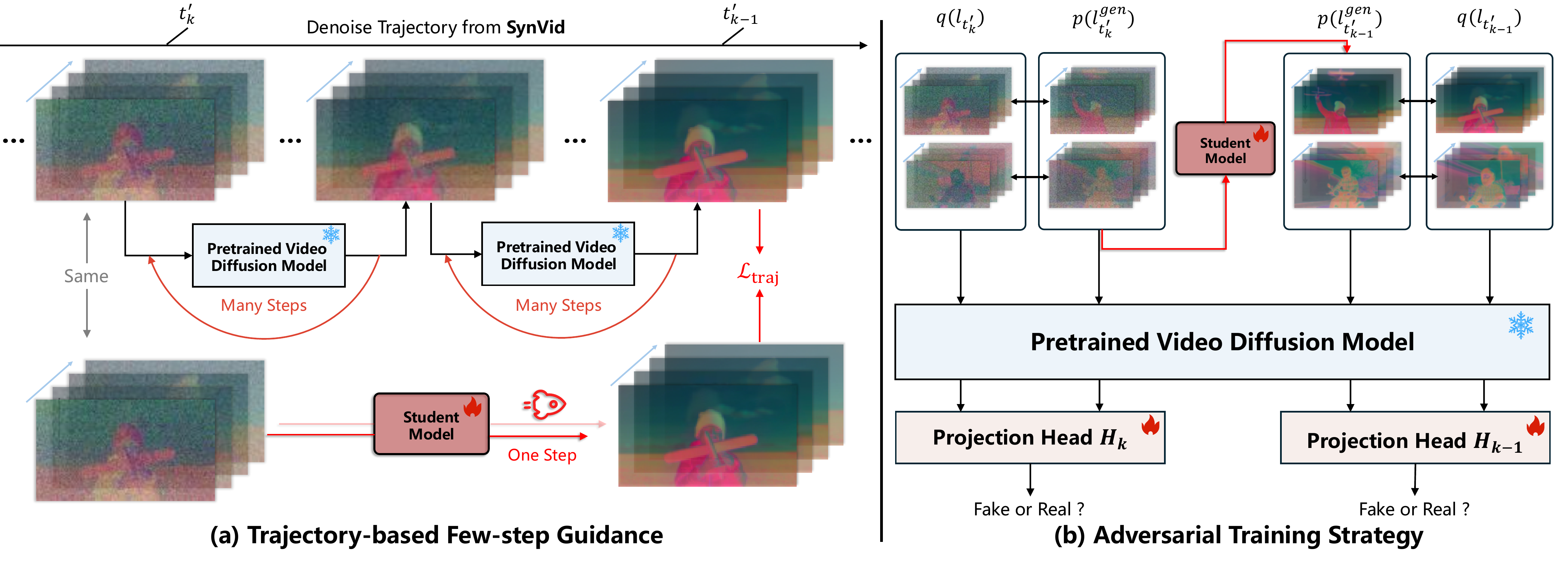}

   \caption{\textbf{Method Overview.} \textbf{(a)} Our method first designs a trajectory-based few-step guidance, which utilizes the key data points from the denoising trajectory to enable the student model to mimic the denoising process of the pretrained video diffusion model with fewer steps. \textbf{(b)} To fully exploit the data distribution at each diffusion timestep captured by our synthetic dataset, we propose an adversarial training strategy to align the output distribution of the student model with that captured by our synthetic dataset.}
   \label{fig:pipeline}
\end{figure*}

Previous diffusion distillation methods can be broadly categorized into two classes: (1) knowledge distillation \cite{salimans2022progressive, frans2024one, yan2024perflow, berthelot2023tract,luo2023latent,li2024t2v,li2024t2v2}, where a student model is trained to mimic the denoising process of a teacher model with fewer inference steps, \eg, mapping $x_{t}$ to $x_{t'}$ or $x_{0}$ using one step, as shown in Fig.~\ref{fig:motivation} a). The start data points $x_{t}$ are obtained through the forward diffusion operation. However, these methods inadvertently distill useless start data points $x_{t}$, which do not lie on the denoising trajectories of the teacher model due to the dataset mismatching or Gaussian noise mismatching, as depicted in Fig.~\ref{fig:motivation} b). When denoising such useless data points using the teacher model, it often produces inaccurate results, as highlighted by the red box in Fig.~\ref{fig:motivation} b). Consequently, it can lead to unreliable guidance for the student model during distillation; (2) distribution distillation \cite{yin2024one, yin2024improved, luo2024one, sauer2024fast, xu2024ufogen, lin2025diffusion}, which leverages diffusion models to compute distribution matching loss or adversarial loss~\cite{wang2022diffusion} to optimize the student model, \eg, feeding $x_{t}^{fake}$ and $x_{t}^{real}$ into the pretrained diffusion model to derive meaningful guidance, as illustrated in Fig.~\ref{fig:motivation} c). However, they may still rely on useless data points $x_{t}^{fake}$ and $x_{t}^{real}$ due to the dataset mismatching or Gaussian noise mismatching, as shown in Fig.~\ref{fig:motivation} d), which leads the diffusion model to provide inaccurate guidance.

We count the frequency of useless data points in relation to the degree of data mismatching and Gaussian noise mismatching in a 1D toy experiment. We define the degree of mismatching between the distillation dataset $p_{d}(x)$ and the training dataset $p(x)$ as follows,
\begin{equation}
    M =\sum_{x_d\in p_{d}(x)}min\{|x_d-x| \ \text{for} \ x \in p(x)\}.
\end{equation}

Fig.~\ref{fig:motivation} e) shows the results. Even without dataset mismatching, \ie, $M=0$, the presence of Gaussian noise mismatching can still produce useless data points. As the degree of dataset mismatching $M$ increases, it becomes evident that the frequency of useless data points grows significantly. Additionally, we conduct distillation experiments followed by \cite{yan2024perflow} at different $M$. The results are illustrated in Fig.~\ref{fig:motivation} f). As $M$ increases, more useless data points are used during distillation, making the teacher model provide incorrect guidance. Consequently, this leads the generated data to deviate from the training dataset $p(x)$ and the distillation dataset $p_d(x)$, demonstrating that useless data points are harmful to the distillation process.

\section{Method}
\label{sec:method}

Our method aims to distill the pretrained video diffusion model \cite{kong2024hunyuanvideo} to reduce the number of inference steps, thereby accelerating video generation. Based on the analysis in Sec.~\ref{sec:analysis}, we avoid using useless data points during the distillation process. Specifically, we first introduce a synthetic video dataset, SynVid, which leverages the teacher model to generate high-quality synthetic videos and denoising trajectories (Sec.~\ref{sec:synvideo}). Subsequently, we propose a trajectory-based few-step guidance that selects key data points from the denoising trajectories and learns the noise-to-video mapping based on these data points, enabling video generation with fewer inference steps (Sec.~\ref{sec:traloss}). Additionally, we introduce an adversarial training strategy that exploits the data distribution at each diffusion timestep captured by SynVid, further enhancing the model's performance (Sec.~\ref{sec:adversarial}). Fig.~\ref{fig:pipeline} illustrates the overview of our method.

\begin{figure}
  \centering
  \setlength{\abovecaptionskip}{0pt}
  \setlength{\belowcaptionskip}{0pt}
  \includegraphics[width=1.\linewidth]{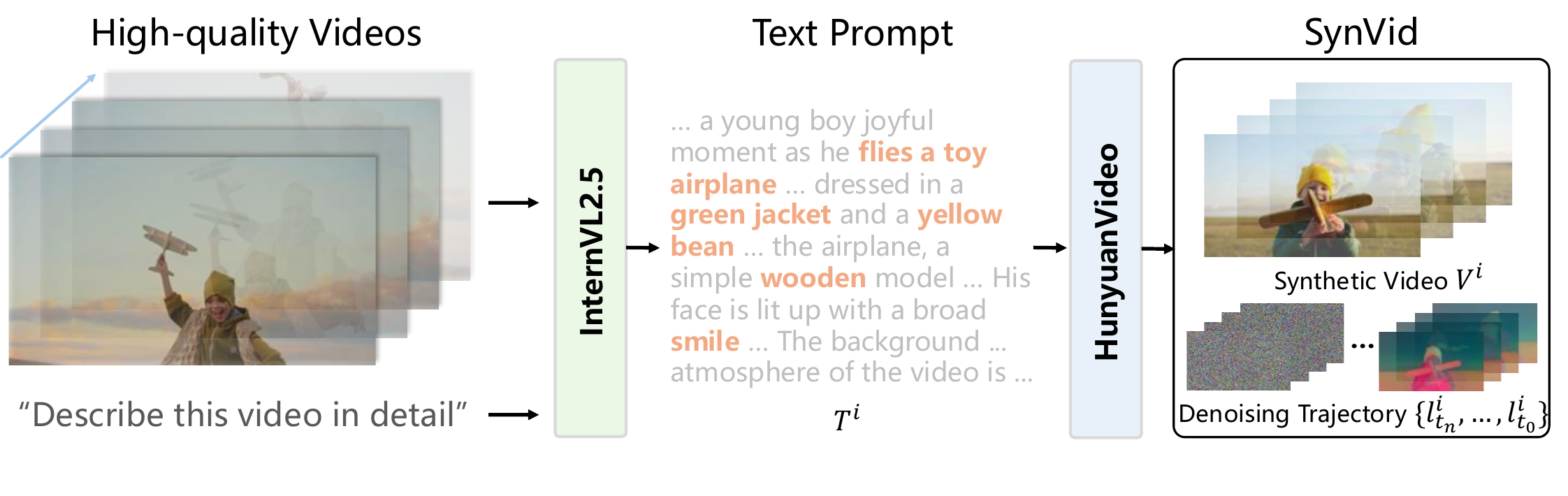}

   \caption{The pipeline of constructing SynVid.}
    \label{fig:synvideo}
    \vspace{-2mm}
\end{figure}

\subsection{SynVid: Synthetic Video Dataset}
\label{sec:synvideo}
We employ HunyuanVideo~\cite{kong2024hunyuanvideo} as our generator to produce synthetic data as shown in Fig.~\ref{fig:synvideo}. HunyuanVideo $v_{\theta}$ is a text-to-video diffusion model utilizing the DiT architecture~\cite{peebles2023scalable}, which operates in the latent space and is trained using the Flow Matching loss, as outlined in Eq.~\ref{training}. We use the official code and settings\footnote{\url{https://github.com/Tencent/HunyuanVideo}} to generate our dataset.

Our SynVid $\mathcal{D}_{\text{syn}}=\{V^i,l_{t_n}^i,l_{t_{n-1}}^i,...,l_{t_0}^i,T^i|i\in[1,N]\}$ comprises high-quality synthetic videos $V_i$, denoising trajectories in latent space $\{l_{t_n}^i,l_{t_{n-1}}^i,...,l_{t_0}^i|t_n=1>...>t_0=0\}$, and their corresponding text prompts $T^i$, where $N=110\text{K}$ represents the number of data, $n=50$ denotes the number of inference steps, and $\{t_j|j\in[0,n]\}$ denote the inference diffusion timesteps. The text prompts $T^i$ are annotated by InternVL2.5 \cite{chen2024expanding} with high-quality videos from the Internet. For a Gaussian noise $l_{t_n}^i$, the denoising trajectory can be solved as follows,
\begin{equation}
     l_{t_{j-1}}^i=l_{t_{j}}^i+(t_{j-1}-t_j)v_{\theta}(l_{t_{j}}^i, t_{j}, T^i),
     \label{denoise_solve}
\end{equation}
the synthetic video $V^i$ can be decoded by VAE using the clean latent $l_{t_0}^i$. For brevity, we omit the text prompt input $T$ in following sections. It is noteworthy that we exclusively utilize the synthetic dataset $\mathcal{D}_{\text{syn}}$ during the distillation process. The data points in $\mathcal{D}_{\text{syn}}$ are intermediate results that lead to the correct output $l_{t_{0}}^i$, making them all valid and meaningful. This significantly aids in efficient distillation and reduces the demand for data.

\begin{figure}
  \centering
  \setlength{\abovecaptionskip}{0pt}
  \setlength{\belowcaptionskip}{0pt}
  \includegraphics[width=1.\linewidth]{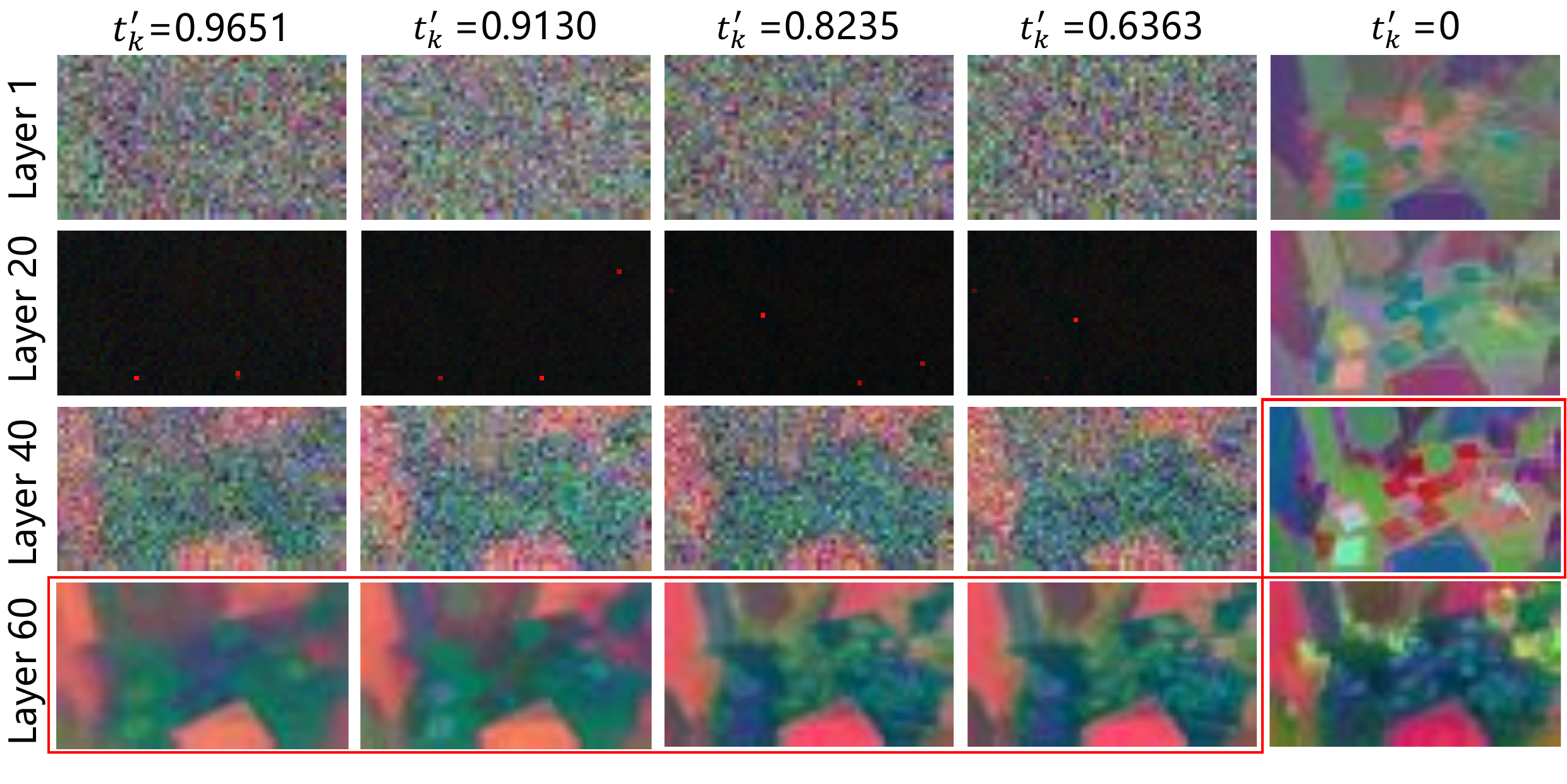}

   \caption{The illustration of features at different layers and diffusion timesteps of our feature extractor. The features within the red box are selected for discrimination.}
   \label{fig:feature}
   \vspace{-2mm}
\end{figure}

\subsection{Trajectory-based Few-step Guidance}
\label{sec:traloss}

Video diffusion models typically require a large number of inference steps to generate videos, which is time-consuming and computationally intensive. To accelerate the generation process, we design a student model $s_{\beta}$ that utilizes denoising trajectories generated by the pretrained video diffusion model, \ie, the teacher model, to learn the noise-to-video mapping with fewer inference steps. 
The student model shares the same architecture as the teacher model and is parameterized by $\beta$, which is initialized using the parameters $\theta$ of the teacher model.

Specifically, we select $m+1$ key diffusion timesteps $\{t'_m=1>t'_{m-1}>...>t'_{0}=0\}$ and obtain their corresponding latents $\{l_{t'_{m}}^{i}, l_{t'_{m-1}}^{i},...,l_{t'_{0}}^{i}\}$ on each denoising trajectory. Then, we propose a trajectory-based loss to learn the denoising process of the teacher model with $m$ steps, 

\begin{equation}
     \mathcal{L}_{\text{traj}}
     = \mathbb{E}_{i, k}\Big\|s_{\beta}(l_{t'_{k+1}}^i,t'_{k+1}) - \frac{l_{t'_{k}}^i-l_{t'_{k+1}}^i}{t'_{k}-t'_{k+1}}) \Big\|^2,
     \label{training_traj}
\end{equation}
where $k\in[0,m-1]$. By learning from these key latents, which construct a shorter path from Gaussian noise to video latent, our student model significantly reducing the number of inference steps. In our experiments, we set $m=5$, which reduces the number of inference steps by a factor of 10 compared to the teacher model, greatly accelerating the generation process.

\begin{algorithm}
    \scriptsize
    \SetKwInOut{KwIn}{Input}
    \SetKwInOut{KwOut}{Output}
    \caption{\label{alg:distillation}AccVideo Distillation Procedure}
    \KwIn{Pretrained video diffusion model $v_\theta$, 
    SynVid $\mathcal{D}_{\text{syn}}$, $m$.}
    \KwOut{Distilled $m$-step student model $s_{\beta}$, projection heads $\{H_{0},...,H_{m-1}\}$.}
    
    \tcp{Initialize student model from pretrained model}
    $\beta \leftarrow \theta$ 
    
    \tcp{Select $m$ key timesteps on denoising trajectory} 
    $t'_m=1>t'_{m-1}>...>t'_{0}=0$
    
    \tcp{Initialize latent queues with $\varnothing$}
    $Q_0, ..., Q_{m} \leftarrow \varnothing$
    
    \While{train}{
        \For{$k \leftarrow m-1$ \KwTo 0}{
            \tcp{Sample key data points}
            $\{l_{t'_{m}}, l_{t'_{m-1}},...,l_{t'_{0}}\}\sim \mathcal{D}_{\text{syn}}$
            
            \text{~}
            
            \tcp{Update student model with $\mathcal{L}_{\text{traj}}$}
            $\mathcal{L}_{\text{traj}} \leftarrow \text{traj}(s_\beta,l_{t'_{k}},l_{t'_{k+1}})\ \ \ $ \tcp{Eq.~\ref{training_traj}}
            ${\beta} \leftarrow \text{Update}({\beta}, \mathcal{L}_{\text{traj}})$

            Sample $z \sim \mathcal{N}(0, \mathbf{I})$
            
            $Q_{m}$.\text{push}($z,l_{t'_{m-1}},...,l_{t'_{0}}$)  \textbf{if} $k=m-1$
            
            \text{~}

            \tcp{Compute latent $l_{t'_k}^{\text{gen}}\sim p_{\beta}(l_{t'_k}^{\text{gen}})$}
            $l_{t'_{k+1}}^{\text{gen}},l_{t'_{m-1}},...,l_{t'_{0}} \leftarrow$ $Q_{k+1}.\text{pop()}$
            
            $l_{t'_{k}}^{\text{gen}}\leftarrow \text{Solver}(s_\beta,l_{t'_{k+1}}^{\text{gen}},t'_{k+1},t'_{k})$ \tcp{Eq.~\ref{denoise_solve}}
            
            $Q_{k}$.\text{push}($l_{t'_{k}}^{\text{gen}},l_{t'_{m-1}},...,l_{t'_{0}}$)

            \text{~} 
            
            \tcp{Update student model and $H_{k}$ with $\mathcal{L}_{\text{adv}}$}
            $\mathcal{L}_{\text{adv}} \leftarrow \text{adv}(l_{t'_{k}}^{\text{gen}},l_{t'_{k}},t'_{k},H_{k},v_\theta)\ \ \ $ \tcp{Eq.~\ref{adversarial}}
            ${\beta} \leftarrow \text{Update}({\beta}, \lambda_{\text{adv}}\mathcal{L}_{\text{adv}})$ \\
            ${H_{k}} \leftarrow \text{Update}(H_{k}, \lambda_{\text{adv}}\mathcal{L}_{\text{adv}})$
            
        }
    }
\end{algorithm}

\subsection{Adversarial Training Strategy}
\label{sec:adversarial}

In addition to the one-to-one mapping between Gaussian noise and video latents, our synthetic dataset also contains many latents $\{l_{t'_k}^{i}|i\in[1,N]\}$ at each diffusion timestep $t'_k$, which implicitly represent the data distribution $q(l_{t'_k})$. To fully unleash the knowledge in our synthetic dataset and enhance the performance of our student model $s_{\beta}$, we propose an adversarial training strategy to minimize adversarial divergence $D_{\text{adv}}\big(q(l_{t'_k})||p(l_{t'_k}^{\text{gen}})\big)$, where $p(l_{t'_k}^{\text{gen}})$ denotes the generated data distribution at diffusion timestep $t'_k$ of our student model $s_{\beta}$. The training objective follows the standard Generative Adversarial Network (GAN)~\cite{goodfellow2014generative}:
\begin{equation}
\begin{aligned}
    \mathcal{L}_{\text{adv}} = & \mathop{\text{min}}\limits_{\beta}\ \mathop{\text{max}}\limits_{\phi}\ \mathbb{E}_{k,q(l_{t'_k}),p(l_{t'_k}^{\text{gen}})} \\
    &\big[\text{log}(D(l_{t'_k},t'_k))+\text{log}(1-D(l_{t'_k}^{\text{gen}},t'_k))\big],
    \label{adversarial}
\end{aligned}
\end{equation}
where $D$ represents the discriminator parameterized by $\phi$. Since the discriminator needs to distinguish noisy latents at different diffusion timesteps, we design our discriminator $D$ to consist of a noise-aware feature extractor and $m$ timestep-aware projection heads $\{H_{0},...,H_{m-1}\}$,
\begin{equation}
\begin{aligned}
    & D(l_{t'_k},t'_k)=H_{k}\big(v_{\theta}(l_{t'_k},t'_k)\big), \\
    &D(l_{t'_k}^{\text{gen}},t'_k)=H_{k}\big(v_{\theta}(l_{t'_k}^{\text{gen}},t'_k)\big),
\end{aligned}
\end{equation}
here, we leverage the frozen teacher model $v_{\theta}$ as the feature extractor inspired by \cite{xu2024ufogen, yin2024improved, lin2024sdxl}. Fig.~\ref{fig:feature} shows the layer-wise behavior at different diffusion timesteps of $v_{\theta}$. When $t'_{k}>0$, the deeper layer concentrate on capturing high-frequency details, so we select the output at the final layer, \ie, the 60th layer for discrimination. When $t'_{k}=0$, we choose the output at the 40th layer for discrimination, which contains fine-grained clues, such as sticky notes on the table. Each projection head $H_{k}$ is a 2D convolutional neural network used to output the label indicating where the latents $l_{t'_k}$ and $l_{t'_k}^{\text{gen}}$ comes from. The projection heads are timestep-aware, effectively handling noisy features at different diffusion timesteps and thereby facilitating adversarial learning.

For latent $l_{t'_k}^{\text{gen}}\sim p_{\beta}(l_{t'_k}^{\text{gen}})$, it can be iteratively solved using $s_{\beta}$, similar to Eq.~\ref{denoise_solve}. However, it is time-consuming, which is not practical implementation. Therefore, we employ $m+1$ queues 
$\{Q_{0},...,Q_{m}\}$, which maintain the latents generated by $s_{\beta}$ at each diffusion timestep. 

It should be noted that our adversarial training strategy avoids the forward diffusion operation commonly used in existing methods~\cite{xu2024ufogen, yin2024improved}, preventing the distillation of useless data points. This ensures that the teacher model provides accurate guidance for the student model. Moreover, the incorporation of our trajectory-based few-step guidance improves the stability of adversarial learning, eliminating the requirement for complex regularization designs~\cite{lin2025diffusion}.

\noindent\textbf{Total Objective.} The training parameters include $\beta$ and the parameters of the projection heads. These parameters are trained using $\mathcal{L}_{\text{traj}}$ and $\lambda_{adv}\mathcal{L}_{\text{adv}}$, where $\lambda_{adv}$ is a hyperparameter. Algorithm~\ref{alg:distillation} shows the distillation procedure.

\begin{figure*}
  \centering
  \setlength{\abovecaptionskip}{0pt}
  \setlength{\belowcaptionskip}{0pt}
  \includegraphics[width=1.\linewidth]{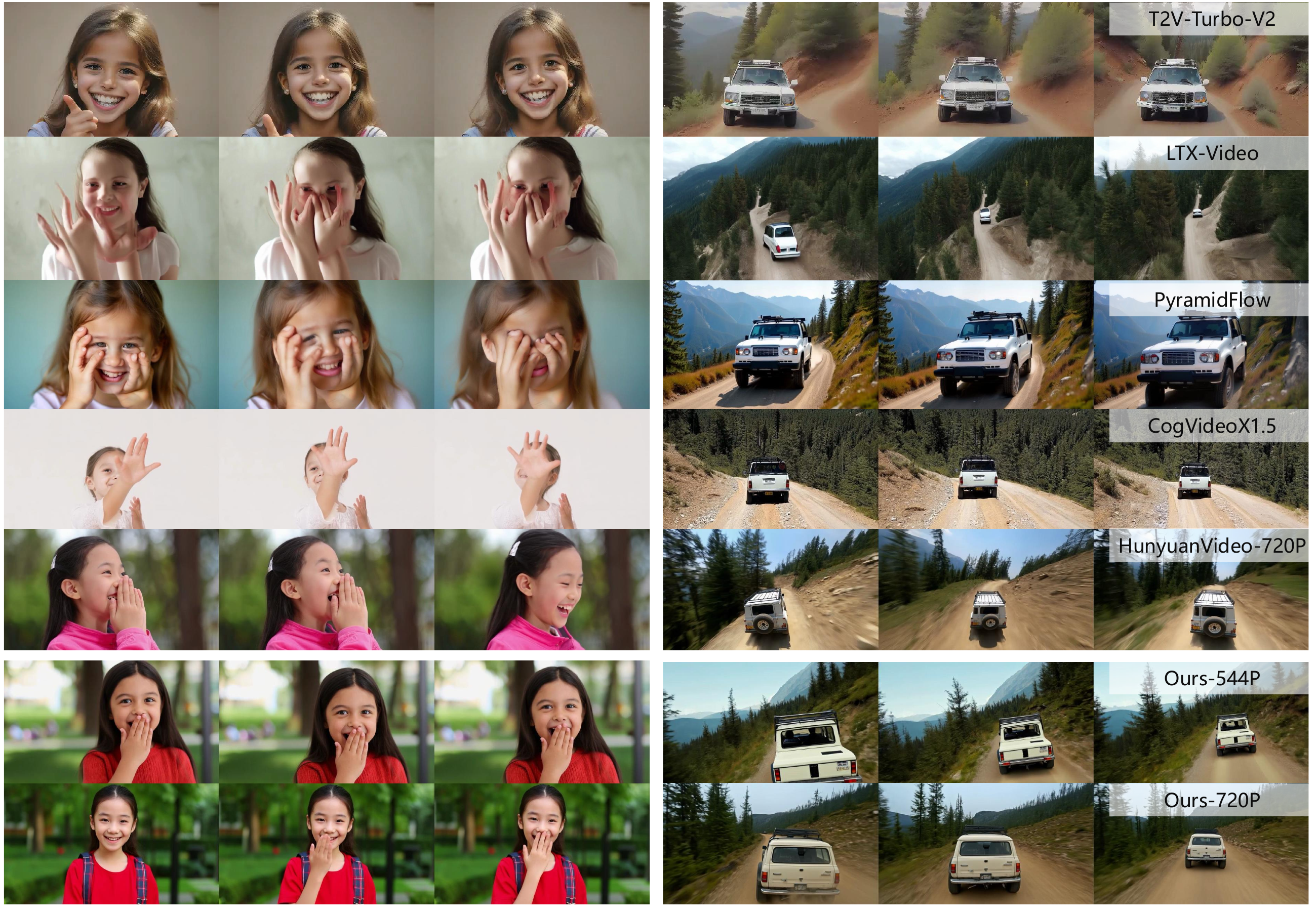}

   \caption{Qualitative results on text-to-video. \textbf{Left:} \textit{a girl raises her left hand to cover her smiling mouth.} \textbf{Right:} \textit{the camera follows behind a white vintage SUV with a black roof rack as it speeds up a steep dirt road surrounded by pine trees on a steep mountain slope.}}
   \label{fig:qualitative}
\end{figure*}
\section{Experiments}
\label{sec:method}

\textbf{Implementation Details.} Our student model and teacher model adopt the HunyuanVideo architecture~\cite{kong2024hunyuanvideo} and are initialized using the officially released checkpoint. The teacher model remains frozen. The student model and projection heads are trained for 1200 iterations using the AdamW optimizer~\cite{loshchilov2017decoupled} with a learning rate of $5\times10^{-6}$  for 12 days, utilizing 8 A100 GPUs and gradient accumulation to achieve a total batch size of 32. The hyperparameter $\lambda_{\text{adv}}$ is set to 0.1. We use SynVid as described in Sec.~\ref{sec:synvideo} as our distillation dataset, where the resolution of the latents and videos is $68\times120\times24$ and $544\times960\times93$, respectively. After distillation, we can generate higher-resolution videos, i.e., $720\times1280\times129$. The inference process in our experiments is conducted on a single A100 GPU.

\noindent\textbf{Evaluation.} Our method is evaluated on VBench~\cite{huang2024vbench}, a comprehensive benchmark that is commonly used to evaluate the video quality. Specifically, VBench evaluates video generation models across 16 dimensions using 946 prompts. Each prompt is sampled five times to reduce randomness.

\noindent\textbf{Baselines.} The baselines include VideoCrafter2, which is a 3DUNet-based model, as well as OpenSora~\cite{zheng2024open}, HunyuanVideo~\cite{kong2024hunyuanvideo}, and CogVideoX~\cite{yang2024cogvideox}, which are DiT-based models. Additionally, we compare with faster models such as T2V-Turbo-V2~\cite{li2024t2v2}, which utilize distillation to accelerate video generation, and LTX-Video~\cite{hacohen2024ltx}, which employs a high-compression VAE to speed up video generation, as well as PyramidFlow~\cite{jin2024pyramidal}, which combines autoregressive and diffusion models for efficient video generation.

\begin{table}
\centering
\caption{Comparison of inference time. Here, we use a single A100 GPU to measure the time required for generating a video, which includes text encoding, VAE decoding, and diffusion time.}
\resizebox{.48\textwidth}{!}{
\begin{tabular}{l|lcc}
\toprule
\textbf{Resolution} & \textbf{Method} & $\textbf{H}\times \textbf{W}\times \textbf{L}$ & \textbf{Inference Time (s)} \\
\midrule

 \multirow{3}{1pt}{Medium} & CogVideoX-5B & $480\times720\times49$ & 219 \\

  & HunyuanVideo-544P & $544\times960\times93$ & 704  \\

\rowcolor{gray!15} & Ours-544P & $544\times960\times93$ & 91 \\
 
\midrule 
 \multirow{3}{1pt}{High}  & CogVideoX1.5-5B & $768\times 1360\times 81$ & 926\\

& HunyuanVideo-720P & $720\times1280\times129$ & 3234\\ 

\rowcolor{gray!15}
& Ours-720P & $720\times1280\times129$ & 380  \\

\bottomrule
\end{tabular}
}
\label{tab:inference_time}
\end{table}

\subsection{Comparisons on Text-to-Video Generation}

\begin{table*}
\centering
\caption{Text-to-video comparisons on VBench~\cite{huang2024vbench}. The results marked with $*$ are tested by us, while the other results are obtained from the official VBench leaderboard. For detailed results across all 16 dimensions, please refer to our supplementary materials.}
\begin{adjustbox}{width=1\textwidth}
\begin{tabular}{l|lccccccccc}
\toprule
\textbf{Resolution} & \textbf{Method} & $\textbf{H}\times \textbf{W}\times \textbf{L}$ & \textbf{Total Score} & \textbf{Quality Score} & \textbf{Semantic Score} & \textbf{Temporal Flickering} & \textbf{Dynamic Degree} & \textbf{Object Class} & \textbf{Color} & \textbf{Spatial Relationship} \\
\midrule

\multirow{2}{1pt}{Low} & VideoCrafter2 & $320\times512\times16$ & 80.44\% & 82.20\% & 73.42\% & 98.41\% & 42.50\% & 92.55\% & 92.92\% & 35.86\% \\

& T2V-Turbo-V2 & $320\times512\times16$ & 83.52\% & 85.13\% & 77.12\% & 97.35\% & 90.00\% & 95.33\% & 92.53\% & 43.32\%  \\

\midrule


\multirow{4}{1pt}{Medium} & LTX-Video & $512\times768\times121$ & 80.00\% & 82.30\% & 70.79\% & 99.34\% & 54.35\% & 83.45\% & 81.45\% & 65.43\% \\

 & CogVideoX-5B & $480\times720\times49$ & 81.61\% & 82.75\% & 77.04\% & 98.66\% & 70.97\% & 85.23\% & 82.81\% & 66.35\%  \\

 & HunyuanVideo-544P$^*$ & $544\times960\times93$ & 82.67\% & 84.43\% & 75.59\% & 99.03\% & 76.38\% & 88.67\% & 92.07\% & 67.39\%  \\ 

\rowcolor{gray!15} & Ours-544P$^*$ & $544\times960\times93$ & 83.26\% & 84.58\% & 77.96\% & 99.18\% & 75.00\% & 92.99\% & 94.11\% & 75.70\% \\ 
 
\midrule 
\multirow{4}{1pt}{High} & PyramidFlow & $768\times 1280\times 121$ & 81.72\% & 84.74\% & 69.62\% & 99.49\% & 64.63\% & 86.67\% & 82.87\% & 59.53\% \\

& OpenSora V1.2 &$720\times 1280\times 204$ & 79.76\% & 81.35\% & 73.39\% & 99.53\% & 42.39\% & 82.22\% &  90.08\% & 68.56\% \\

 & CogVideoX1.5-5B & $768\times 1360\times 81$ & 82.17\% & 82.78\% & 79.76\% & 98.88\% & 50.93\% & 87.47\% & 87.55\% & 80.25\% \\

& HunyuanVideo-720P & $720\times1280\times129$ & 83.24\% & 85.09\% & 75.82\% & 99.44\% & 70.83\% & 86.10\% & 91.60\% & 68.68\% \\ 

\rowcolor{gray!15}
& Ours-720P$^*$ & $720\times1280\times129$ & 82.77\% & 84.38\% & 76.34\% & 99.01\% & 75.55\% & 89.71\% & 94.61\% & 71.74\%  \\

\bottomrule
\end{tabular}
\end{adjustbox}
\label{tab:vbench}
\end{table*}

\noindent\textbf{Inference Time.} Tab.~\ref{tab:inference_time} presents the comparison of the inference time. Our model requires only 5 inference steps to generate videos, it achieves a 7.7-8.5$\times$ improvement in generation speed compared to the teacher model. Even compared to models with half the model size of ours, \eg, CogVideoX-5B and CogVideoX1.5-5B~\cite{yang2024cogvideox}, our model is still 2.7$\times$ faster.

\noindent\textbf{Quantitative Evaluation.} The evaluations on VBench are presented in Tab.~\ref{tab:vbench}. T2V-Turbo-V2~\cite{li2024t2v2} achieves leading results in total score, primarily due to their highest performance in dynamic degree (improving by at least 13\%). However, this may be attributed to its imperfect temporal consistency in local areas, as evidenced by its poor performance in temporal flickering. Moreover, it struggles to generate high-resolution videos, whereas our model retains this ability. In addition to T2V-Turbo-V2~\cite{li2024t2v2} and our teacher model, HunyuanVideo~\cite{kong2024hunyuanvideo}, our method surpasses all the other compared baselines at the same resolution level. Compared to our teacher model, we achieve comparable performance at high resolution and better performance at medium resolution. In particular, we achieve exceptional performance in color and spatial relationship, indicating that our model can effectively interpret text prompts and generate videos coherent with the text. This not only demonstrates the effectiveness of our method but also highlights that our synthetic dataset contains high-quality text prompts and data points.

\noindent\textbf{Qualitative Evaluation.} Fig.~\ref{fig:qualitative} presents the qualitative results. Compared to T2V-Turbo~\cite{li2024t2v2}, LTX-Video~\cite{hacohen2024ltx}, and PyramidFlow~\cite{jin2024pyramidal}, our method generates videos with fewer artifacts, such as the hands of the little girl. Compared to CogVideoX1.5~\cite{yang2024cogvideox}, we produce higher-fidelity videos with better backgrounds. Compared to HunyuanVideo~\cite{kong2024hunyuanvideo}, our method better aligns with the text prompts, such as the vintage SUV. For more qualitative results, please refer to our supplementary materials.

\begin{figure}
  \centering
  \setlength{\abovecaptionskip}{0pt}
  \setlength{\belowcaptionskip}{0pt}
  \includegraphics[width=1.\linewidth]{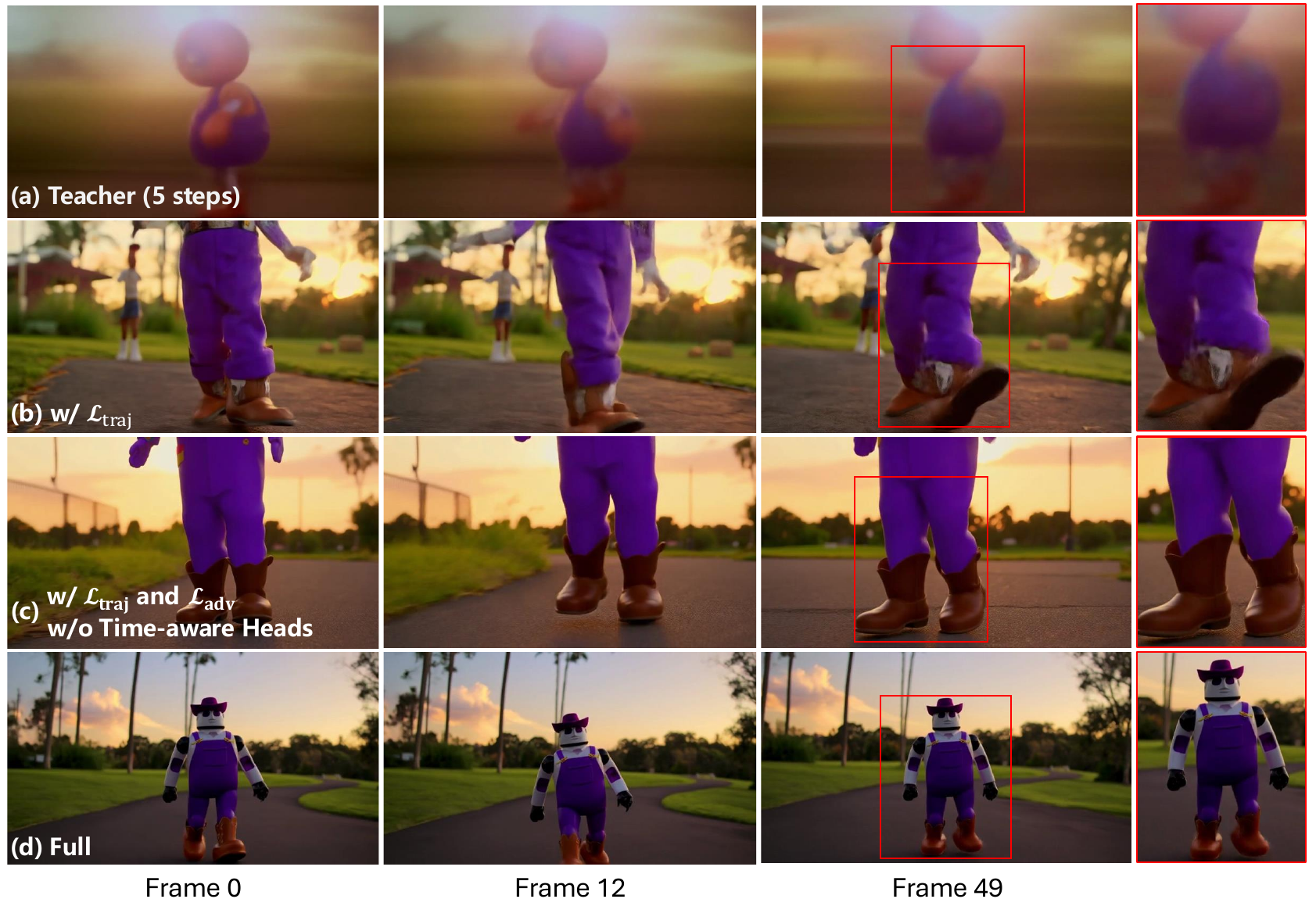}

   \caption{Qualitative ablation study results. Please zoom in for details. Prompt: \textit{a toy robot wearing purple overalls and cowboy boots taking a pleasant stroll in Johannesburg South Africa during a beautiful sunset.}}
   \label{fig:ablation}
\end{figure}

\subsection{Ablation Studies}

\noindent\textbf{Trajectory-based Few-step Guidance.} The trajectory-based few-step guidance is designed to guide the student model to learn the denoising trajectories of the teacher model with fewer steps. As illustrated in Fig.~\ref{fig:ablation} (a) and (b), we observe that the teacher model struggles to generate clear videos with fewer inference steps, while distilling with our trajectory-based loss is capable of producing clearer videos in just 5 steps. We greatly accelerate video generation by reducing the number of inference steps,.

\noindent\textbf{Adversarial Training Strategy.} As shown in Fig.~\ref{fig:ablation} (b) and (c), although the trajectory-based few-step guidance accelerates video generation, the generated videos contain artifacts with unnatural motions. In contrast, when the adversarial training strategy is employed, we can generate higher-quality videos. This further demonstrates its effectiveness.

\noindent\textbf{Timestep-Aware Projection Heads.} We conduct an ablation study, which only uses one projection head to discriminate the extracted features. As illustrated in Fig.~\ref{fig:ablation} (c) and (d), the generated videos better align with the text prompts and produce more coherent videos when using our timestep-aware projection heads. We hypothesize that when a single projection head is used, it may struggle to distinguish features from different diffusion timesteps, \ie, different noise levels. In contrast, the timestep-aware projection heads can more effectively learn variance present in these features.

\section{Conclusion}
\label{sec:conclusion}

In this paper, we first analyze the challenges faced by existing diffusion distillation methods, which are caused by the use of useless data points. Based on this insight, we propose a novel efficient distillation method to accelerate video diffusion models. Specifically, we first construct a synthetic video dataset, SynVid, which contains valid and meaningful data points for distillation. Then, we introduce a trajectory-based few-step guidance that enables the student model to generate videos in just 5 steps, significantly accelerating the generation speed compared to the teacher model. To further enhance the video quality, we design an adversarial training strategy that leverages the data distribution captured by our synthetic video dataset. Our model achieves 8.5$\times$ improvements in generation speed compared to the teacher model while maintaining comparable performance.
{
    \small
    \bibliographystyle{ieeenat_fullname}
    \bibliography{main}

\begin{thebibliography}{66}
\providecommand{\natexlab}[1]{#1}
\providecommand{\url}[1]{\texttt{#1}}
\expandafter\ifx\csname urlstyle\endcsname\relax
  \providecommand{\doi}[1]{doi: #1}\else
  \providecommand{\doi}{doi: \begingroup \urlstyle{rm}\Url}\fi

\bibitem[Agarwal et~al.(2025)Agarwal, Ali, Bala, Balaji, Barker, Cai, Chattopadhyay, Chen, Cui, Ding, et~al.]{agarwal2025cosmos}
Niket Agarwal, Arslan Ali, Maciej Bala, Yogesh Balaji, Erik Barker, Tiffany Cai, Prithvijit Chattopadhyay, Yongxin Chen, Yin Cui, Yifan Ding, et~al.
\newblock Cosmos world foundation model platform for physical ai.
\newblock \emph{arXiv preprint arXiv:2501.03575}, 2025.

\bibitem[Berthelot et~al.(2023)Berthelot, Autef, Lin, Yap, Zhai, Hu, Zheng, Talbott, and Gu]{berthelot2023tract}
David Berthelot, Arnaud Autef, Jierui Lin, Dian~Ang Yap, Shuangfei Zhai, Siyuan Hu, Daniel Zheng, Walter Talbott, and Eric Gu.
\newblock Tract: Denoising diffusion models with transitive closure time-distillation.
\newblock \emph{arXiv preprint arXiv:2303.04248}, 2023.

\bibitem[Blattmann et~al.(2023)Blattmann, Rombach, Ling, Dockhorn, Kim, Fidler, and Kreis]{blattmann2023align}
Andreas Blattmann, Robin Rombach, Huan Ling, Tim Dockhorn, Seung~Wook Kim, Sanja Fidler, and Karsten Kreis.
\newblock Align your latents: High-resolution video synthesis with latent diffusion models.
\newblock In \emph{CVPR}, pages 22563--22575, 2023.

\bibitem[Chen et~al.(2024{\natexlab{a}})Chen, Zhang, Cun, Xia, Wang, Weng, and Shan]{chen2024videocrafter2}
Haoxin Chen, Yong Zhang, Xiaodong Cun, Menghan Xia, Xintao Wang, Chao Weng, and Ying Shan.
\newblock Videocrafter2: Overcoming data limitations for high-quality video diffusion models.
\newblock In \emph{CVPR}, pages 7310--7320, 2024{\natexlab{a}}.

\bibitem[Chen et~al.(2025)Chen, Cai, Chen, Xie, Yang, Tang, Li, Lu, and Han]{chen2024deep}
Junyu Chen, Han Cai, Junsong Chen, Enze Xie, Shang Yang, Haotian Tang, Muyang Li, Yao Lu, and Song Han.
\newblock Deep compression autoencoder for efficient high-resolution diffusion models.
\newblock In \emph{ICLR}, 2025.

\bibitem[Chen et~al.(2024{\natexlab{b}})Chen, Xu, Ren, Cong, He, Xie, Sinha, Luo, Xiang, and Perez-Rua]{chen2023gentron}
Shoufa Chen, Mengmeng Xu, Jiawei Ren, Yuren Cong, Sen He, Yanping Xie, Animesh Sinha, Ping Luo, Tao Xiang, and Juan-Manuel Perez-Rua.
\newblock Gentron: Delving deep into diffusion transformers for image and video generation.
\newblock In \emph{CVPR}, 2024{\natexlab{b}}.

\bibitem[Chen et~al.(2024{\natexlab{c}})Chen, Wang, Cao, Liu, Gao, Cui, Zhu, Ye, Tian, Liu, et~al.]{chen2024expanding}
Zhe Chen, Weiyun Wang, Yue Cao, Yangzhou Liu, Zhangwei Gao, Erfei Cui, Jinguo Zhu, Shenglong Ye, Hao Tian, Zhaoyang Liu, et~al.
\newblock Expanding performance boundaries of open-source multimodal models with model, data, and test-time scaling.
\newblock \emph{arXiv preprint arXiv:2412.05271}, 2024{\natexlab{c}}.

\bibitem[Dao and Gu(2024)]{dao2024transformers}
Tri Dao and Albert Gu.
\newblock Transformers are ssms: Generalized models and efficient algorithms through structured state space duality.
\newblock In \emph{ICML}, 2024.

\bibitem[Frans et~al.(2025)Frans, Hafner, Levine, and Abbeel]{frans2024one}
Kevin Frans, Danijar Hafner, Sergey Levine, and Pieter Abbeel.
\newblock One step diffusion via shortcut models.
\newblock In \emph{ICLR}, 2025.

\bibitem[Goodfellow et~al.(2024)Goodfellow, Pouget-Abadie, Mirza, Xu, Warde-Farley, Ozair, Courville, and Bengio]{goodfellow2014generative}
Ian Goodfellow, Jean Pouget-Abadie, Mehdi Mirza, Bing Xu, David Warde-Farley, Sherjil Ozair, Aaron Courville, and Yoshua Bengio.
\newblock Generative adversarial nets.
\newblock In \emph{NIPS}, 2024.

\bibitem[Guo et~al.(2023)Guo, Yang, Rao, Liang, Wang, Qiao, Agrawala, Lin, and Dai]{guo2023animatediff}
Yuwei Guo, Ceyuan Yang, Anyi Rao, Zhengyang Liang, Yaohui Wang, Yu Qiao, Maneesh Agrawala, Dahua Lin, and Bo Dai.
\newblock Animatediff: Animate your personalized text-to-image diffusion models without specific tuning.
\newblock \emph{arXiv preprint arXiv:2307.04725}, 2023.

\bibitem[HaCohen et~al.(2024)HaCohen, Chiprut, Brazowski, Shalem, Moshe, Richardson, Levin, Shiran, Zabari, Gordon, et~al.]{hacohen2024ltx}
Yoav HaCohen, Nisan Chiprut, Benny Brazowski, Daniel Shalem, Dudu Moshe, Eitan Richardson, Eran Levin, Guy Shiran, Nir Zabari, Ori Gordon, et~al.
\newblock Ltx-video: Realtime video latent diffusion.
\newblock \emph{arXiv preprint arXiv:2501.00103}, 2024.

\bibitem[He et~al.(2016)He, Zhang, Ren, and Sun]{he2016deep}
Kaiming He, Xiangyu Zhang, Shaoqing Ren, and Jian Sun.
\newblock Deep residual learning for image recognition.
\newblock In \emph{CVPR}, pages 770--778, 2016.

\bibitem[He et~al.(2022)He, Yang, Zhang, Shan, and Chen]{he2022latent}
Yingqing He, Tianyu Yang, Yong Zhang, Ying Shan, and Qifeng Chen.
\newblock Latent video diffusion models for high-fidelity long video generation.
\newblock \emph{arXiv preprint arXiv:2211.13221}, 2022.

\bibitem[Ho et~al.(2020)Ho, Jain, and Abbeel]{ho2020denoising}
Jonathan Ho, Ajay Jain, and Pieter Abbeel.
\newblock Denoising diffusion probabilistic models.
\newblock \emph{NIPS}, pages 6840--6851, 2020.

\bibitem[Ho et~al.(2022{\natexlab{a}})Ho, Chan, Saharia, Whang, Gao, Gritsenko, Kingma, Poole, Norouzi, Fleet, et~al.]{ho2022imagen}
Jonathan Ho, William Chan, Chitwan Saharia, Jay Whang, Ruiqi Gao, Alexey Gritsenko, Diederik~P Kingma, Ben Poole, Mohammad Norouzi, David~J Fleet, et~al.
\newblock Imagen video: High definition video generation with diffusion models.
\newblock \emph{arXiv preprint arXiv:2210.02303}, 2022{\natexlab{a}}.

\bibitem[Ho et~al.(2022{\natexlab{b}})Ho, Salimans, Gritsenko, Chan, Norouzi, and Fleet]{ho2022video}
Jonathan Ho, Tim Salimans, Alexey Gritsenko, William Chan, Mohammad Norouzi, and David~J Fleet.
\newblock Video diffusion models.
\newblock In \emph{NIPS}, pages 8633--8646, 2022{\natexlab{b}}.

\bibitem[Hu et~al.(2024)Hu, Gao, Zhang, Sun, Zhang, and Bo]{hu2024animate}
Li Hu, Xin Gao, Peng Zhang, Ke Sun, Bang Zhang, and Liefeng Bo.
\newblock Animate anyone: Consistent and controllable image-to-video synthesis for character animation.
\newblock In \emph{CVPR}, pages 8153--8163, 2024.

\bibitem[Huang et~al.(2024)Huang, He, Yu, Zhang, Si, Jiang, Zhang, Wu, Jin, Chanpaisit, et~al.]{huang2024vbench}
Ziqi Huang, Yinan He, Jiashuo Yu, Fan Zhang, Chenyang Si, Yuming Jiang, Yuanhan Zhang, Tianxing Wu, Qingyang Jin, Nattapol Chanpaisit, et~al.
\newblock Vbench: Comprehensive benchmark suite for video generative models.
\newblock In \emph{CVPR}, pages 21807--21818, 2024.

\bibitem[Jin et~al.(2025)Jin, Sun, Li, Xu, Jiang, Zhuang, Huang, Song, Mu, and Lin]{jin2024pyramidal}
Yang Jin, Zhicheng Sun, Ningyuan Li, Kun Xu, Hao Jiang, Nan Zhuang, Quzhe Huang, Yang Song, Yadong Mu, and Zhouchen Lin.
\newblock Pyramidal flow matching for efficient video generative modeling.
\newblock In \emph{ICLR}, 2025.

\bibitem[Karras et~al.(2022)Karras, Aittala, Aila, and Laine]{karras2022elucidating}
Tero Karras, Miika Aittala, Timo Aila, and Samuli Laine.
\newblock Elucidating the design space of diffusion-based generative models.
\newblock \emph{NIPS}, 35:\penalty0 26565--26577, 2022.

\bibitem[Kingma(2014)]{kingma2013auto}
Diederik~P Kingma.
\newblock Auto-encoding variational bayes.
\newblock In \emph{ICLR}, 2014.

\bibitem[Kong et~al.(2024)Kong, Tian, Zhang, Min, Dai, Zhou, Xiong, Li, Wu, Zhang, et~al.]{kong2024hunyuanvideo}
Weijie Kong, Qi Tian, Zijian Zhang, Rox Min, Zuozhuo Dai, Jin Zhou, Jiangfeng Xiong, Xin Li, Bo Wu, Jianwei Zhang, et~al.
\newblock Hunyuanvideo: A systematic framework for large video generative models.
\newblock \emph{arXiv preprint arXiv:2412.03603}, 2024.

\bibitem[Kuaishou(2024)]{kling}
Kuaishou.
\newblock Kling, 2024.
\newblock \url{https://klingai.kuaishou.com}.

\bibitem[Li et~al.(2024)Li, Feng, Fu, Wang, Basu, Chen, and Wang]{li2024t2v}
Jiachen Li, Weixi Feng, Tsu-Jui Fu, Xinyi Wang, Sugato Basu, Wenhu Chen, and William~Yang Wang.
\newblock T2v-turbo: Breaking the quality bottleneck of video consistency model with mixed reward feedback.
\newblock In \emph{NIPS}, 2024.

\bibitem[Li et~al.(2025)Li, Long, Zheng, Gao, Piramuthu, Chen, and Wang]{li2024t2v2}
Jiachen Li, Qian Long, Jian Zheng, Xiaofeng Gao, Robinson Piramuthu, Wenhu Chen, and William~Yang Wang.
\newblock T2v-turbo-v2: Enhancing video generation model post-training through data, reward, and conditional guidance design.
\newblock In \emph{ICLR}, 2025.

\bibitem[Lin et~al.(2025{\natexlab{a}})Lin, Jiang, Yang, Zheng, and Liang]{lin2025omnihuman}
Gaojie Lin, Jianwen Jiang, Jiaqi Yang, Zerong Zheng, and Chao Liang.
\newblock Omnihuman-1: Rethinking the scaling-up of one-stage conditioned human animation models.
\newblock \emph{arXiv preprint arXiv:2502.01061}, 2025{\natexlab{a}}.

\bibitem[Lin and Yang(2024)]{lin2024animatediff}
Shanchuan Lin and Xiao Yang.
\newblock Animatediff-lightning: Cross-model diffusion distillation.
\newblock \emph{arXiv preprint arXiv:2403.12706}, 2024.

\bibitem[Lin et~al.(2024)Lin, Wang, and Yang]{lin2024sdxl}
Shanchuan Lin, Anran Wang, and Xiao Yang.
\newblock Sdxl-lightning: Progressive adversarial diffusion distillation.
\newblock \emph{arXiv preprint arXiv:2402.13929}, 2024.

\bibitem[Lin et~al.(2025{\natexlab{b}})Lin, Xia, Ren, Yang, Xiao, and Jiang]{lin2025diffusion}
Shanchuan Lin, Xin Xia, Yuxi Ren, Ceyuan Yang, Xuefeng Xiao, and Lu Jiang.
\newblock Diffusion adversarial post-training for one-step video generation.
\newblock \emph{arXiv preprint arXiv:2501.08316}, 2025{\natexlab{b}}.

\bibitem[Lipman et~al.(2023)Lipman, Chen, Ben-Hamu, Nickel, and Le]{lipman2022flow}
Yaron Lipman, Ricky~TQ Chen, Heli Ben-Hamu, Maximilian Nickel, and Matt Le.
\newblock Flow matching for generative modeling.
\newblock In \emph{ICLR}, 2023.

\bibitem[Liu et~al.(2023{\natexlab{a}})Liu, Gong, and Liu]{liu2022flow}
Xingchao Liu, Chengyue Gong, and Qiang Liu.
\newblock Flow straight and fast: Learning to generate and transfer data with rectified flow.
\newblock In \emph{ICLR}, 2023{\natexlab{a}}.

\bibitem[Liu et~al.(2023{\natexlab{b}})Liu, Zhang, Ma, Peng, et~al.]{liu2023instaflow}
Xingchao Liu, Xiwen Zhang, Jianzhu Ma, Jian Peng, et~al.
\newblock Instaflow: One step is enough for high-quality diffusion-based text-to-image generation.
\newblock In \emph{ICLR}, 2023{\natexlab{b}}.

\bibitem[Loshchilov and Hutter(2019)]{loshchilov2017decoupled}
Ilya Loshchilov and Frank Hutter.
\newblock Decoupled weight decay regularization.
\newblock In \emph{ICLR}, 2019.

\bibitem[Luhman and Luhman(2021)]{luhman2021knowledge}
Eric Luhman and Troy Luhman.
\newblock Knowledge distillation in iterative generative models for improved sampling speed.
\newblock \emph{arXiv preprint arXiv:2101.02388}, 2021.

\bibitem[Luo et~al.(2023)Luo, Tan, Huang, Li, and Zhao]{luo2023latent}
Simian Luo, Yiqin Tan, Longbo Huang, Jian Li, and Hang Zhao.
\newblock Latent consistency models: Synthesizing high-resolution images with few-step inference.
\newblock \emph{arXiv preprint arXiv:2310.04378}, 2023.

\bibitem[Luo et~al.(2024)Luo, Huang, Geng, Kolter, and Qi]{luo2024one}
Weijian Luo, Zemin Huang, Zhengyang Geng, J~Zico Kolter, and Guo-jun Qi.
\newblock One-step diffusion distillation through score implicit matching.
\newblock In \emph{NIPS}, 2024.

\bibitem[Ma et~al.(2024{\natexlab{a}})Ma, Wang, Jia, Chen, Liu, Li, Chen, and Qiao]{ma2024latte}
Xin Ma, Yaohui Wang, Gengyun Jia, Xinyuan Chen, Ziwei Liu, Yuan-Fang Li, Cunjian Chen, and Yu Qiao.
\newblock Latte: Latent diffusion transformer for video generation.
\newblock \emph{arXiv preprint arXiv:2401.03048}, 2024{\natexlab{a}}.

\bibitem[Ma et~al.(2024{\natexlab{b}})Ma, He, Cun, Wang, Chen, Li, and Chen]{ma2024follow}
Yue Ma, Yingqing He, Xiaodong Cun, Xintao Wang, Siran Chen, Xiu Li, and Qifeng Chen.
\newblock Follow your pose: Pose-guided text-to-video generation using pose-free videos.
\newblock In \emph{AAAI}, pages 4117--4125, 2024{\natexlab{b}}.

\bibitem[Menapace et~al.(2024)Menapace, Siarohin, Skorokhodov, Deyneka, Chen, Kag, Fang, Stoliar, Ricci, Ren, et~al.]{menapace2024snap}
Willi Menapace, Aliaksandr Siarohin, Ivan Skorokhodov, Ekaterina Deyneka, Tsai-Shien Chen, Anil Kag, Yuwei Fang, Aleksei Stoliar, Elisa Ricci, Jian Ren, et~al.
\newblock Snap video: Scaled spatiotemporal transformers for text-to-video synthesis.
\newblock In \emph{CVPR}, pages 7038--7048, 2024.

\bibitem[OpenAI(2024)]{sora}
OpenAI.
\newblock Sora, 2024.
\newblock \url{https://openai.com/sora}.

\bibitem[Peebles and Xie(2023)]{peebles2023scalable}
William Peebles and Saining Xie.
\newblock Scalable diffusion models with transformers.
\newblock In \emph{ICCV}, pages 4195--4205, 2023.

\bibitem[Poole et~al.(2023)Poole, Jain, Barron, and Mildenhall]{poole2022dreamfusion}
Ben Poole, Ajay Jain, Jonathan~T Barron, and Ben Mildenhall.
\newblock Dreamfusion: Text-to-3d using 2d diffusion.
\newblock In \emph{ICLR}, 2023.

\bibitem[Rombach et~al.(2022)Rombach, Blattmann, Lorenz, Esser, and Ommer]{rombach2022high}
Robin Rombach, Andreas Blattmann, Dominik Lorenz, Patrick Esser, and Bj{\"o}rn Ommer.
\newblock High-resolution image synthesis with latent diffusion models.
\newblock In \emph{CVPR}, pages 10684--10695, 2022.

\bibitem[Ronneberger et~al.(2015)Ronneberger, Fischer, and Brox]{ronneberger2015u}
Olaf Ronneberger, Philipp Fischer, and Thomas Brox.
\newblock U-net: Convolutional networks for biomedical image segmentation.
\newblock In \emph{MICCAI}, pages 234--241, 2015.

\bibitem[Salimans and Ho(2022)]{salimans2022progressive}
Tim Salimans and Jonathan Ho.
\newblock Progressive distillation for fast sampling of diffusion models.
\newblock In \emph{ICLR}, 2022.

\bibitem[Sauer et~al.(2024{\natexlab{a}})Sauer, Boesel, Dockhorn, Blattmann, Esser, and Rombach]{sauer2024fast}
Axel Sauer, Frederic Boesel, Tim Dockhorn, Andreas Blattmann, Patrick Esser, and Robin Rombach.
\newblock Fast high-resolution image synthesis with latent adversarial diffusion distillation.
\newblock In \emph{SIGGRAPH Asia}, pages 1--11, 2024{\natexlab{a}}.

\bibitem[Sauer et~al.(2024{\natexlab{b}})Sauer, Lorenz, Blattmann, and Rombach]{sauer2024adversarial}
Axel Sauer, Dominik Lorenz, Andreas Blattmann, and Robin Rombach.
\newblock Adversarial diffusion distillation.
\newblock In \emph{ECCV}, pages 87--103, 2024{\natexlab{b}}.

\bibitem[Song et~al.(2023)Song, Dhariwal, Chen, and Sutskever]{song2023consistency}
Yang Song, Prafulla Dhariwal, Mark Chen, and Ilya Sutskever.
\newblock Consistency models.
\newblock 2023.

\bibitem[Valevski et~al.(2024)Valevski, Leviathan, Arar, and Fruchter]{valevski2024diffusion}
Dani Valevski, Yaniv Leviathan, Moab Arar, and Shlomi Fruchter.
\newblock Diffusion models are real-time game engines.
\newblock \emph{arXiv preprint arXiv:2408.14837}, 2024.

\bibitem[Wang et~al.(2024{\natexlab{a}})Wang, Ma, Liu, Hou, Xu, Wang, Juefei-Xu, Luo, Zhang, Hou, et~al.]{wang2024lingen}
Hongjie Wang, Chih-Yao Ma, Yen-Cheng Liu, Ji Hou, Tao Xu, Jialiang Wang, Felix Juefei-Xu, Yaqiao Luo, Peizhao Zhang, Tingbo Hou, et~al.
\newblock Lingen: Towards high-resolution minute-length text-to-video generation with linear computational complexity.
\newblock \emph{arXiv preprint arXiv:2412.09856}, 2024{\natexlab{a}}.

\bibitem[Wang et~al.(2023{\natexlab{a}})Wang, Yuan, Chen, Zhang, Wang, and Zhang]{wang2023modelscope}
Jiuniu Wang, Hangjie Yuan, Dayou Chen, Yingya Zhang, Xiang Wang, and Shiwei Zhang.
\newblock Modelscope text-to-video technical report.
\newblock \emph{arXiv preprint arXiv:2308.06571}, 2023{\natexlab{a}}.

\bibitem[Wang et~al.(2024{\natexlab{b}})Wang, Chen, Ma, Zhou, Huang, Wang, Yang, He, Yu, Yang, et~al.]{wang2024lavie}
Yaohui Wang, Xinyuan Chen, Xin Ma, Shangchen Zhou, Ziqi Huang, Yi Wang, Ceyuan Yang, Yinan He, Jiashuo Yu, Peiqing Yang, et~al.
\newblock Lavie: High-quality video generation with cascaded latent diffusion models.
\newblock \emph{International Journal of Computer Vision}, pages 1--20, 2024{\natexlab{b}}.

\bibitem[Wang et~al.(2023{\natexlab{b}})Wang, Zheng, He, Chen, and Zhou]{wang2022diffusion}
Zhendong Wang, Huangjie Zheng, Pengcheng He, Weizhu Chen, and Mingyuan Zhou.
\newblock Diffusion-gan: Training gans with diffusion.
\newblock In \emph{ICLR}, 2023{\natexlab{b}}.

\bibitem[Wang et~al.(2024{\natexlab{c}})Wang, Lu, Wang, Bao, Li, Su, and Zhu]{wang2024prolificdreamer}
Zhengyi Wang, Cheng Lu, Yikai Wang, Fan Bao, Chongxuan Li, Hang Su, and Jun Zhu.
\newblock Prolificdreamer: High-fidelity and diverse text-to-3d generation with variational score distillation.
\newblock In \emph{NIPS}, 2024{\natexlab{c}}.

\bibitem[Xie et~al.(2025)Xie, Chen, Chen, Cai, Tang, Lin, Zhang, Li, Zhu, Lu, et~al.]{xie2024sana}
Enze Xie, Junsong Chen, Junyu Chen, Han Cai, Haotian Tang, Yujun Lin, Zhekai Zhang, Muyang Li, Ligeng Zhu, Yao Lu, et~al.
\newblock Sana: Efficient high-resolution image synthesis with linear diffusion transformers.
\newblock In \emph{ICLR}, 2025.

\bibitem[Xing et~al.(2024)Xing, Feng, Chen, Dai, Hu, Xu, Wu, and Jiang]{xing2024survey}
Zhen Xing, Qijun Feng, Haoran Chen, Qi Dai, Han Hu, Hang Xu, Zuxuan Wu, and Yu-Gang Jiang.
\newblock A survey on video diffusion models.
\newblock \emph{ACM Computing Surveys}, 57\penalty0 (2):\penalty0 1--42, 2024.

\bibitem[Xu et~al.(2024)Xu, Zhao, Xiao, and Hou]{xu2024ufogen}
Yanwu Xu, Yang Zhao, Zhisheng Xiao, and Tingbo Hou.
\newblock Ufogen: You forward once large scale text-to-image generation via diffusion gans.
\newblock In \emph{CVPR}, pages 8196--8206, 2024.

\bibitem[Yan et~al.(2024)Yan, Liu, Pan, Liew, Liu, and Feng]{yan2024perflow}
Hanshu Yan, Xingchao Liu, Jiachun Pan, Jun~Hao Liew, Qiang Liu, and Jiashi Feng.
\newblock Perflow: Piecewise rectified flow as universal plug-and-play accelerator.
\newblock In \emph{NIPS}, 2024.

\bibitem[Yang et~al.(2024)Yang, Li, Fang, Chen, Yu, Fu, Yang, and Ye]{yang2024playable}
Mingyu Yang, Junyou Li, Zhongbin Fang, Sheng Chen, Yangbin Yu, Qiang Fu, Wei Yang, and Deheng Ye.
\newblock Playable game generation.
\newblock \emph{arXiv preprint arXiv:2412.00887}, 2024.

\bibitem[Yang et~al.(2025)Yang, Teng, Zheng, Ding, Huang, Xu, Yang, Hong, Zhang, Feng, et~al.]{yang2024cogvideox}
Zhuoyi Yang, Jiayan Teng, Wendi Zheng, Ming Ding, Shiyu Huang, Jiazheng Xu, Yuanming Yang, Wenyi Hong, Xiaohan Zhang, Guanyu Feng, et~al.
\newblock Cogvideox: Text-to-video diffusion models with an expert transformer.
\newblock In \emph{ICLR}, 2025.

\bibitem[Yin et~al.(2024{\natexlab{a}})Yin, Gharbi, Park, Zhang, Shechtman, Durand, and Freeman]{yin2024improved}
Tianwei Yin, Micha{\"e}l Gharbi, Taesung Park, Richard Zhang, Eli Shechtman, Fredo Durand, and William~T Freeman.
\newblock Improved distribution matching distillation for fast image synthesis.
\newblock In \emph{NIPS}, 2024{\natexlab{a}}.

\bibitem[Yin et~al.(2024{\natexlab{b}})Yin, Gharbi, Zhang, Shechtman, Durand, Freeman, and Park]{yin2024one}
Tianwei Yin, Micha{\"e}l Gharbi, Richard Zhang, Eli Shechtman, Fredo Durand, William~T Freeman, and Taesung Park.
\newblock One-step diffusion with distribution matching distillation.
\newblock In \emph{CVPR}, pages 6613--6623, 2024{\natexlab{b}}.

\bibitem[Yin et~al.(2024{\natexlab{c}})Yin, Zhang, Zhang, Freeman, Durand, Shechtman, and Huang]{yin2024slow}
Tianwei Yin, Qiang Zhang, Richard Zhang, William~T Freeman, Fredo Durand, Eli Shechtman, and Xun Huang.
\newblock From slow bidirectional to fast causal video generators.
\newblock \emph{arXiv preprint arXiv:2412.07772}, 2024{\natexlab{c}}.

\bibitem[Zheng et~al.(2024)Zheng, Peng, Yang, Shen, Li, Liu, Zhou, Li, and You]{zheng2024open}
Zangwei Zheng, Xiangyu Peng, Tianji Yang, Chenhui Shen, Shenggui Li, Hongxin Liu, Yukun Zhou, Tianyi Li, and Yang You.
\newblock Open-sora: Democratizing efficient video production for all.
\newblock \emph{arXiv preprint arXiv:2412.20404}, 2024.

\bibitem[Zhou et~al.(2022)Zhou, Wang, Yan, Lv, Zhu, and Feng]{zhou2022magicvideo}
Daquan Zhou, Weimin Wang, Hanshu Yan, Weiwei Lv, Yizhe Zhu, and Jiashi Feng.
\newblock Magicvideo: Efficient video generation with latent diffusion models.
\newblock \emph{arXiv preprint arXiv:2211.11018}, 2022.

\end{thebibliography}
}

\clearpage
\maketitlesupplementary

\renewcommand{\thesection}{\Alph{section}}
\setcounter{section}{0}

\begin{figure}
  \centering
  \setlength{\abovecaptionskip}{0pt}
  \setlength{\belowcaptionskip}{0pt}
  \includegraphics[width=1.\linewidth]{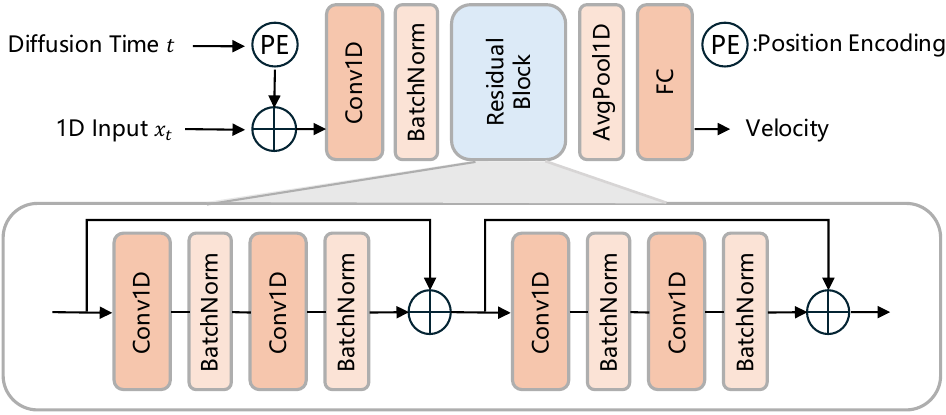}

   \caption{Model architecture for the 1D toy experiment.}
   \label{fig:model_archi}
   \vspace{-2mm}
\end{figure}

\section{Details about the 1D Toy Experiment}

In Sec. 3.2, we conduct a 1D toy experiment to analyze existing diffusion distillation methods. Specifically, we set the training data as $\{-3,3\}$. The training loss is shown in Eq. (1). We use ResNet~\cite{he2016deep} to construct our model, as illustrated in Fig.~\ref{fig:model_archi}. The model is trained using the Adam optimizer for 10000 iterations with a learning rate of $10^{-4}$. Although our dataset consists of only two data points, we achieve a batch size of 2048 by repeating the data. 

\section{SynVid}

\begin{figure}
  \centering
  \setlength{\abovecaptionskip}{0pt}
  \setlength{\belowcaptionskip}{0pt}
  \includegraphics[width=1.\linewidth]{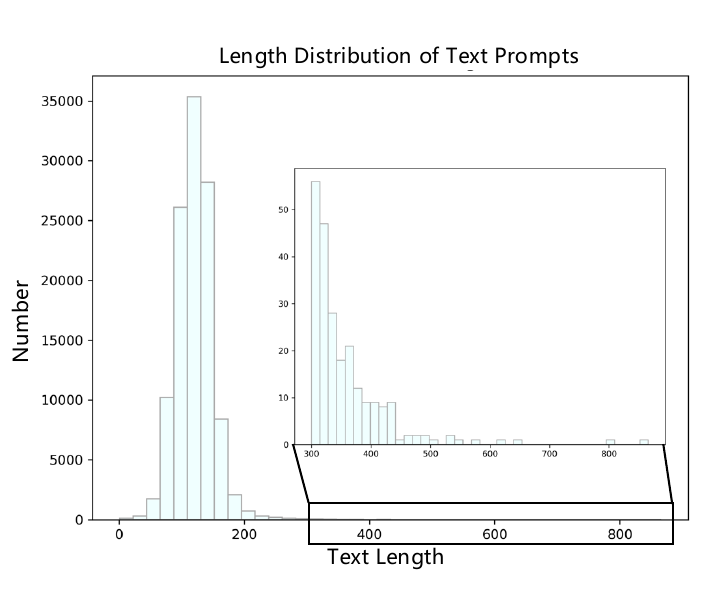}

   \caption{The length distribution of text prompts.}
   \label{fig:text_prompt}
   \vspace{-2mm}
\end{figure}

\begin{figure}
  \centering
  \setlength{\abovecaptionskip}{0pt}
  \setlength{\belowcaptionskip}{0pt}
  \includegraphics[width=1.\linewidth]{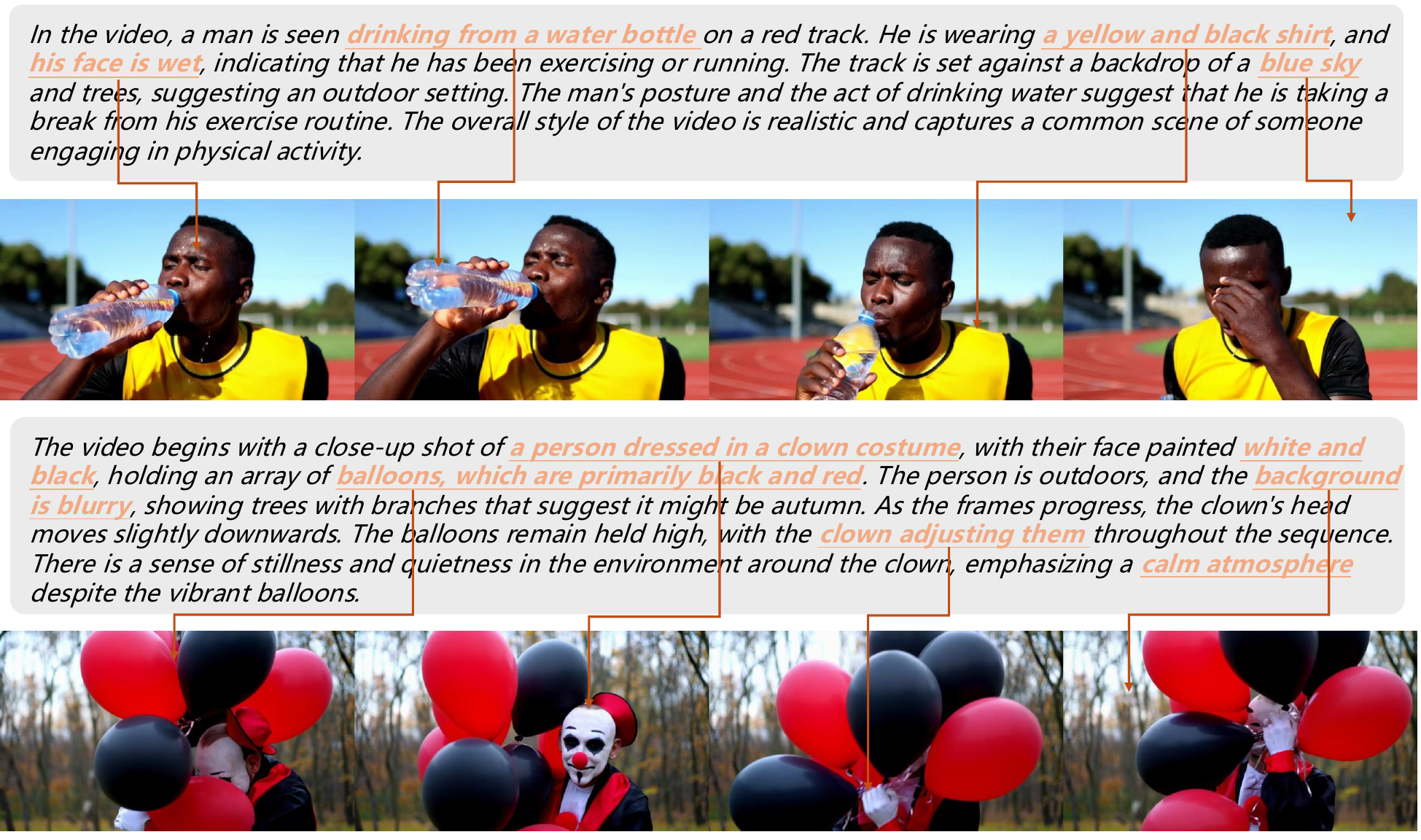}

   \caption{The illustration of our synthetic video dataset, SynVid.}
   \label{fig:synvid_ill}
   \vspace{-2mm}
\end{figure}

Recent methods have demonstrated that fine-grained text prompts play a crucial role in video generation. Inspired by this, we leverage a multimodal large language model (MLLM), i.e., InternVL2.5-8B~\cite{chen2024expanding}, to annotate real videos and obtain high-quality text prompts. Fig.~\ref{fig:text_prompt} shows the length distribution of text prompts, most text lengths are concentrated between 100 and 150, demonstrating that the text prompts provide rich semantic information. As depicted in Fig.~\ref{fig:synvid_ill}, the text prompts accurately describe the people and objects present in the videos, along with details about them, as well as dynamic motions. Moreover, they include descriptions of the atmosphere and background, which greatly aids in generating high-quality videos. The high-quality synthetic dataset facilitates the model distillation.

To promote community development and support future research, SynVid includes videos of different resolutions, \ie, medium and high resolution, with detailed information provided in Tab.~\ref{tab:detail_vid}.

\section{More Results}

\textbf{Full Comparisons on VBench.} We evaluate our method on VBench benchmark. The full comparisons across 16 dimensions on VBench are shown in Tab.~\ref{tab:vbench_full_1} and Tab.~\ref{tab:vbench_full_2}. 

\begin{figure*}
  \centering
  \setlength{\abovecaptionskip}{0pt}
  \setlength{\belowcaptionskip}{0pt}
  \includegraphics[width=1.\linewidth]{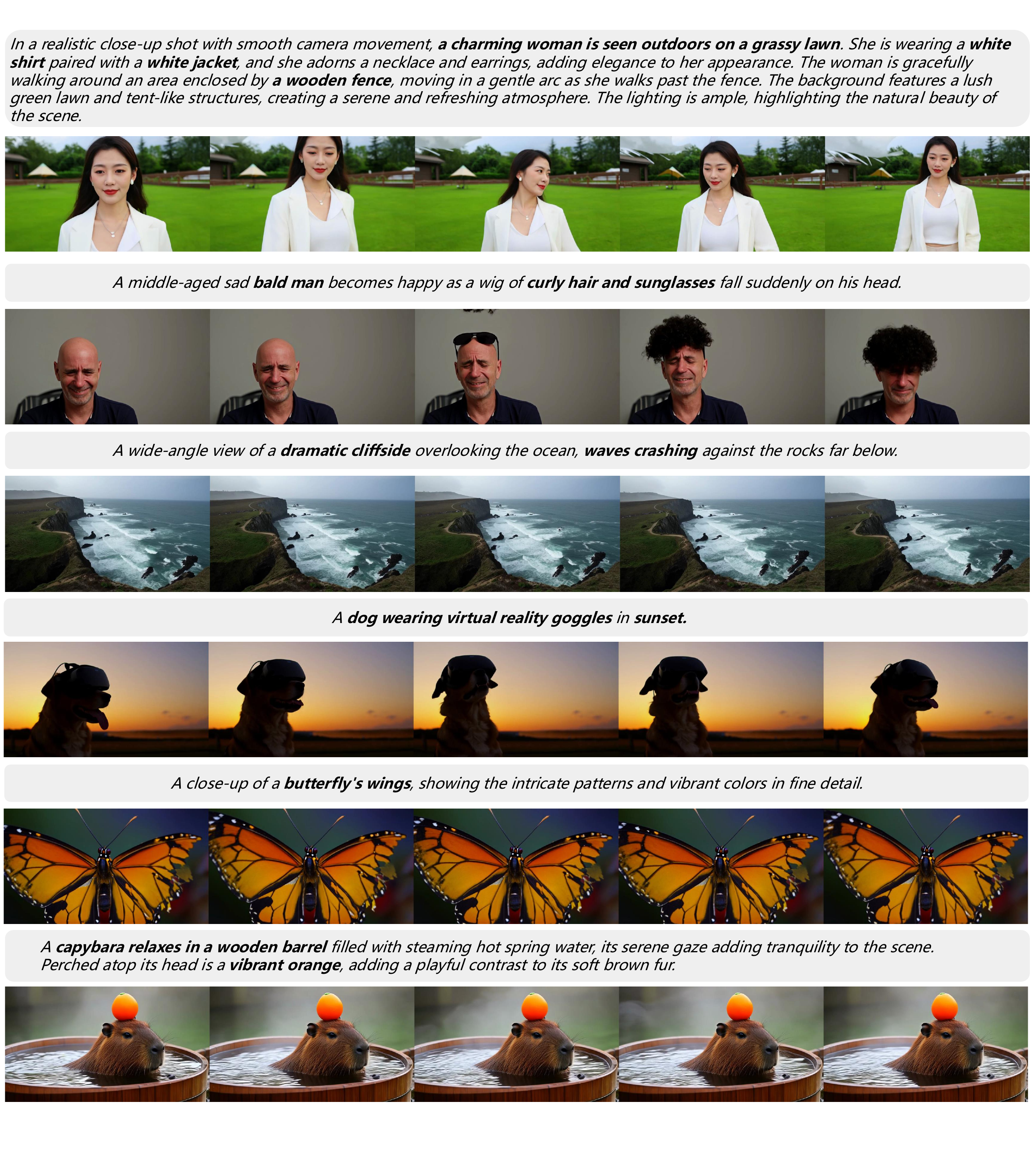}
    \vspace{-10mm}
   \caption{Qualitative results on text-to-video with high resolution, \ie, 720$\times$1280$\times$129.}
   \label{fig:supp_720p}
\end{figure*}

\begin{figure*}
  \centering
  \setlength{\abovecaptionskip}{0pt}
  \setlength{\belowcaptionskip}{0pt}
  \includegraphics[width=1.\linewidth]{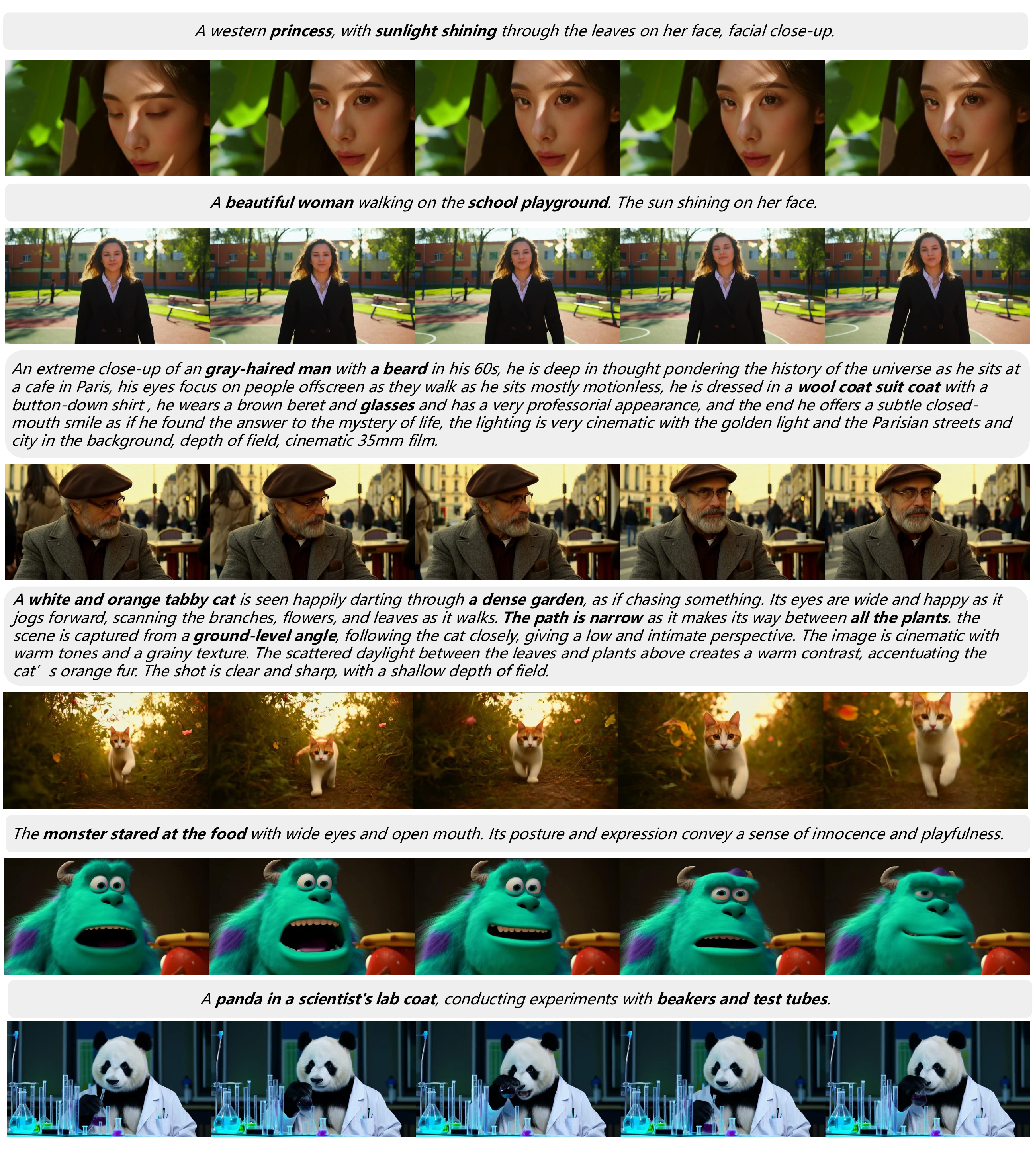}
   \caption{Qualitative results on text-to-video with medium resolution, \ie, 544$\times$960$\times$93.}
   \label{fig:supp_544p}
\end{figure*}

\noindent\textbf{More Qualitative Results.} Fig.~\ref{fig:supp_720p} and Fig.~\ref{fig:supp_544p} showcase the results of our method in generating videos at high and medium resolutions, respectively.

\begin{table}
\centering
\caption{Details resolutions about our synthetic dataset, SynVid.}
\resizebox{.48\textwidth}{!}{
\begin{tabular}{cccc}
\toprule
 Resolution & $\textbf{H}\times \textbf{W}\times \textbf{L}$ (Video) & $\textbf{H}\times \textbf{W}\times \textbf{L}$ (Latents) & Number \\
\midrule

High & 720$\times$1280$\times$129 & 90$\times$160$\times$33 & 8303 \\
Medium & 544$\times$960$\times$93 & 68$\times$120$\times$24 & 105996 \\

\bottomrule
\end{tabular}
}
\label{tab:detail_vid}
\end{table}

\begin{table*}
\centering
\caption{Full comparison on VBench using all 16 metrics.}
\vspace{-2mm}
\begin{adjustbox}{width=1\textwidth}
\begin{tabular}{l|lcccccccccccccccccccc}
\toprule
\textbf{Resolution} & \textbf{Method} & $\textbf{H}\times \textbf{W}\times \textbf{L}$ & \makecell{\textbf{Total} \\ \textbf{Score}} & \makecell{\textbf{Quality} \\ \textbf{Score}} & \makecell{\textbf{Semantic} \\ \textbf{Score}} & \makecell{\textbf{Subject} \\ \textbf{Consistency}} & \makecell{\textbf{Background} \\ \textbf{Consistency}} & \makecell{\textbf{Temporal} \\ \textbf{Flickering}} & \makecell{\textbf{Motion} \\ \textbf{Smoothness}} & \makecell{\textbf{Dynamic} \\ \textbf{Degree}} & \makecell{\textbf{Aesthetic} \\ \textbf{Quality}} \\
\midrule

\multirow{2}{1pt}{Low} & VideoCrafter2 & $320\times512\times16$ & 80.44\% & 82.20\% & 73.42\% & 96.85\% & 98.22\% & 98.41\% & 97.73\% & 42.50\% & 63.13\%\\

& T2V-Turbo-V2 & $320\times512\times16$ & 83.52\% & 85.13\% & 77.12\% & 95.50\% & 96.71\% & 97.35\% & 97.07\% & 90.00\% & 62.61\% \\

\midrule

\multirow{4}{1pt}{Medium} & LTX-Video & $512\times768\times121$ & 80.00\% & 82.30\% & 70.79\% & 96.56\% & 97.20\% & 99.34\% & 98.96\% & 54.35\% & 59.81\% \\

 & CogVideoX & $480\times720\times49$ & 81.61\% & 82.75\% & 77.04\% & 96.23\% & 96.52\% & 98.66\% & 96.92\% & 70.97\% & 61.98\% \\

 & HunyuanVideo-544P & $544\times960\times93$ & 82.67\% & 84.43\% & 75.59\% & 94.33\% & 97.13\% & 99.03\% & 98.64\% & 76.38\% & 61.97\% \\ 

\rowcolor{gray!15}
 & Ours-544P & $544\times960\times93$ & 83.26\% & 84.58\% & 77.96\% & 94.46\% & 97.45\% & 99.18\% & 98.79\% & 75.00\% & 62.08\% \\ 
 
\midrule 
\multirow{5}{1pt}{High} & PyramidFlow & $768\times 1280\times 121$ & 81.72\% & 84.74\% & 69.62\% & 96.95\% & 98.06\% & 99.49\% & 99.12\% & 64.63\% & 63.26\% \\

& OpenSora V1.2 &$720\times 1280\times 204$ & 79.76\% & 81.35\% & 73.39\% & 96.75\% & 97.61\% & 99.53\% & 98.50\% & 42.39\% & 56.85\% \\

 & CogVideoX1.5 & $768\times 1360\times 81$ & 82.17\% & 82.78\% & 79.76\% & 96.87\% & 97.35\% & 98.88\% & 98.31\% & 50.93\% & 62.79\% \\

& HunyuanVideo-720P & $720\times1280\times129$ & 83.24\% & 85.09\% & 75.82\% & 97.37\% & 97.76\% & 99.44\% & 98.99\% & 70.83\% & 60.36\% \\ 

\rowcolor{gray!15}
& Ours-720P & $720\times1280\times129$ & 82.77\% & 84.38\% & 76.34\% & 94.20\% & 96.87\% & 99.01\% & 98.90\% & 75.55\% & 60.61\%   \\

\bottomrule
\end{tabular}
\end{adjustbox}
\label{tab:vbench_full_1}
\end{table*}

\begin{table*}
\centering
\caption{Full comparison on VBench using all 16 metrics.}
\vspace{-2mm}
\begin{adjustbox}{width=1\textwidth}
\begin{tabular}{l|lcccccccccccccccccccc}
\toprule
\textbf{Resolution} & \textbf{Method} & $\textbf{H}\times \textbf{W}\times \textbf{L}$ & \makecell{\textbf{Imaging} \\ \textbf{Quality}} & \makecell{\textbf{Object} \\ \textbf{Class}} & \makecell{\textbf{Multiple} \\ \textbf{Objects}} & \makecell{\textbf{Human} \\ \textbf{Action}} & \textbf{Color} & \makecell{\textbf{Spatial} \\ \textbf{Relationship}} & \textbf{Scene} & \makecell{\textbf{Appearance} \\ \textbf{Style}} & \makecell{\textbf{Temporal} \\ \textbf{Style}} & \makecell{\textbf{Overall} \\ \textbf{Consistency}} \\
\midrule

\multirow{2}{1pt}{Low} & VideoCrafter2 & $320\times512\times16$ & 67.22\% & 92.55\% & 40.66\% & 95.00\% & 92.92\% & 35.86\% & 55.29\% & 25.13\% & 25.84\% & 28.23\% \\

& T2V-Turbo-V2 & $320\times512\times16$ & 71.78\% & 95.33\% & 61.49\% & 96.20\% & 92.53\% & 43.32\% & 56.40\% & 24.17\% & 27.06\% & 28.26\% \\

\midrule

\multirow{4}{1pt}{Medium} & LTX-Video & $512\times768\times121$ & 60.28\% & 83.45\% & 45.43\% & 92.80\% & 81.45\% & 65.43\% & 51.07\% & 21.47\% & 22.62\% & 25.19\% \\

 & CogVideoX & $480\times720\times49$ & 62.90\% & 85.23\% & 62.11\% & 99.40\% & 82.81\% & 66.35\% & 53.20\% & 24.91\% & 25.38\% & 27.59\% \\

 & HunyuanVideo-544P & $544\times960\times93$ & 65.57\% & 88.67\% & 67.69\% & 94.80\% & 92.07\% & 67.39\% & 51.23\% & 19.43\% & 23.96\% & 26.80\% \\ 

\rowcolor{gray!15}
 & Ours-544P & $544\times960\times93$ & 65.64\% & 92.99\% & 67.33\% & 95.60\% & 94.11\% & 75.70\% & 54.72\% & 19.87\% & 23.71\% & 27.21\% \\ 
 
\midrule 
\multirow{5}{1pt}{High} & PyramidFlow & $768\times 1280\times 121$ & 65.01\% & 86.67\% & 50.71\% & 85.60\% & 82.87\% & 59.53\% & 43.20\% & 20.91\% & 23.09\% & 26.23\% \\

& OpenSora V1.2 &$720\times 1280\times 204$ & 63.34\% & 82.22\% & 51.83\% & 91.20\% & 90.08\% & 68.56\% & 42.44\% & 23.95\% & 24.54\% & 26.85\% \\

 & CogVideoX1.5 & $768\times 1360\times 81$ & 65.02\% & 87.47\% & 69.65\% & 97.20\% & 87.55\% & 80.25\% & 52.91\% & 24.89\% & 25.19\% & 27.30\% \\

& HunyuanVideo-720P & $720\times1280\times129$ & 67.56\% & 86.10\% & 68.55\% & 94.40\% & 91.60\% & 68.68\% & 53.88\% & 19.80\% & 23.89\% & 26.44\% \\ 

\rowcolor{gray!15}
& Ours-720P & $720\times1280\times129$ & 66.70\% & 89.71\% & 66.35\% & 94.00\% & 94.61\% & 71.74\% & 50.75\% & 20.16\% & 23.44\% & 26.92\% \\

\bottomrule
\end{tabular}
\end{adjustbox}
\label{tab:vbench_full_2}
\end{table*}

\section{Limitations and Future Work}

Our method focuses on accelerating video diffusion models by distillation technique, which reduces the number of inference steps. However, generation speed is also influenced by VAE, which are used to encode and decode videos, making the acceleration of the VAE another promising direction for exploration~\cite{hacohen2024ltx, chen2024deep}. Additionally, the DiT architecture contains many transformer blocks, and accelerating these transformer blocks is another area that warrants further investigation~\cite{wang2024lingen, xie2024sana}.

\end{document}